\documentclass[twoside,11pt]{article}

\usepackage{jmlr2eopt}

\usepackage{amsmath,bm,soul}  

\usepackage{bbm}

\usepackage{multirow}

\usepackage{enumerate}

\usepackage{wrapfig}
\usepackage{subcaption}

\newcommand*\pct{\scalebox{.9}{\%}}
\newcommand*{\QEDA}{\hfill\BlackBox}

\usepackage{algorithm}
\usepackage{algpseudocode}

\usepackage{cleveref}

\def\E{{\mathbb E}}

\def\Pr{{\mathbb{P}}}

\def\Re{\mathbb{R}}

\def\I{\mathbb{I}}

\def\hat{\widehat}

\def \F{\mathcal{F}}
\def \A{\mathcal{A}}
\def \R{\mathcal{R}}

\def\N{{\mathcal N}}
\def\F{{\mathcal F}}
\def\D{{\mathcal D}}

\def\Re{{\mathbb R}}

\newcommand{\supp}{\mathrm{Supp}}
\newcommand{\bfx}{\bm{x}}
\newcommand{\bfw}{\bm{w}}
\newcommand{\bfu}{\bm{u}}
\newcommand{\bfz}{\bm{z}}

\newcommand{\e}{\bm{e}}

\algtext*{EndWhile}
\algtext*{EndIf}
\algdef{SE}[DOWHILE]{Do}{doWhile}{\algorithmicdo}[1]{\algorithmicwhile\ #1}%
\algnewcommand\INPUT{\item[\textbf{Input:}]}%
\algnewcommand\OUTPUT{\item[\textbf{Output:}]}%
\def\F{{\mathcal F}}

\usepackage{tikz}
\usetikzlibrary{shapes,decorations,arrows,calc,arrows.meta,fit,positioning}
\tikzset{
	state/.style ={ellipse, draw, minimum width = 0.7 cm},
}
\usetikzlibrary{decorations.pathreplacing}

\allowdisplaybreaks

\usepackage[flushleft]{threeparttable}

\jmlrheading{}{}{}{}{}{}{}

\ShortHeadings{Unbiased Subdata Selection for Fair Classification}{Qing Ye and Weijun Xie}
\firstpageno{1}

\begin{document}

\title{Unbiased Subdata Selection for Fair Classification: \\A Unified Framework and Scalable Algorithms}

\author{\name Qing Ye \email yqing1@vt.edu \\
       \addr Department of Industrial and Systems Engineering\\
       Virginia Tech\\
       Blacksburg, VA 24061, USA
       \AND
       \name Weijun Xie \email wxie@vt.edu\\
       \addr Department of Industrial and Systems Engineering\\
       Virginia Tech\\
	   Blacksburg, VA 24061, USA}
%
%
	   

\editor{}
\maketitle

\begin{abstract}
As an important problem in modern data analytics, classification has witnessed varieties of applications from different domains. Different from the conventional classification approaches, fair classification concerns the issues of unintentional biases against the sensitive features (e.g., gender, race). Due to high nonconvexity of fairness measures, existing methods are often unable to model exact fairness, which can cause inferior fair classification outcomes. This paper fills the gap by developing a novel unified framework to jointly optimize accuracy and fairness. The proposed framework is versatile and can incorporate different fairness measures studied in literature precisely as well as can be applicable to many classifiers including deep classification models. Specifically, in this paper, we first prove Fisher consistency of the proposed framework. We then show that many classification models within this framework can be recast as mixed-integer convex programs, which can be solved effectively by off-the-shelf solvers when the instance sizes are moderate and can be used as benchmarks to compare the efficiency of approximation algorithms. We prove that in the proposed framework, when the classification outcomes are known, the resulting problem, termed ``unbiased subdata selection,” is strongly polynomial-solvable and can be used to enhance the classification fairness by selecting more representative data points. This motivates us to develop an iterative refining strategy (IRS) to solve the large-scale instances, where we improve the classification accuracy and conduct the unbiased subdata selection in an alternating fashion. We study the convergence property of IRS and derive its approximation bound. More broadly, this framework can be further leveraged to improve classification models with unbalanced data by taking $F_1$ score into consideration. Finally, we numerically demonstrate that the proposed framework can consistently yield better fair classification outcomes than existing methods.
\end{abstract}


\section{Introduction}
As an important problem in modern data analytics, classification has witnessed varieties of applications including diagnosis \citep{de2016machine}, face recognition \citep{naseem2010linear}, text categorization \citep{zhang2001text}, and microarray gene expression \citep{pirooznia2008comparative}. Different from the conventional classification approaches, fair classification concerns the discrimination against the protected groups when performing the classification tasks.  Recently, fairness in machine learning has attracted much attention since more and more evidences have shown that the traditional methods might cause biases against sensitive features, such as gender, race, or ethnicity. One example of machine learning biases is in criminal risk assessment using the COMPAS recidivism algorithm \citep{ProPublica}. 
In 2016, \citet{ProPublica} analyzed the COMPAS recidivism algorithm and concluded that black defendants were twice as likely as white defendants to be misclassified as being at a high risk of recidivism. 
To prevent such discrimination, many notions of machine learning fairness, in particular, classification fairness, have been established. For example, one popular fair classification measure is overall misclassification rate (OMR) \citep{zafar2019fairness}, which accounts for the misclassification rates of disparate mistreatment among different groups; 
another fair classification measure is demographic parity \citep{menon2018cost}, which measures statistical independence between the classification outcomes and the sensitive feature. 
Due to nonconvexity, existing methods are often unable to model exact fairness, which can cause inferior fair classification outcomes. To foster the impact of machine learning on society progressively, this paper paves a generic way to incorporate the exact fairness measures to improve classification fairness and computational efficiency.

\subsection{Relevant Literature}
In this subsection, we present an overview of fairness measures and existing methods for fair classification. 

\noindent\textbf{Fairness Measures:}
According to \cite{verma2018fairness}, fairness measures for classification problems can be classified as statistical measures, similarity-based measures, and casual reasoning. Statistical measures such as group fairness, predictive parity, and test-fairness are developed based on confusion matrix with their different emphases on actual outcome, predicted outcome, or predicted probabilities. 
Similarity-based measures such as causal discrimination and fairness through unawareness incorporate insensitive features using data similarity. Casual reasoning such as counter-factual fairness and no unresolved discrimination define fairness measures for causal graphs that are used to build fair classifiers \citep{kilbertus2017avoiding}. This paper mainly focuses on the fundamental statistical measures, where disparate treatment, disparate impact, and disparate mistreatment are the common fairness notions \citep{zafar2019fairness,taskesen2020distributionally}. 
Specifically, a classifier achieves no disparate treatment if the prediction results are independent of the sensitive feature, while a classifier does not suffer from the disparate impact if its proportional prediction outcome of a specific label is the same to the different groups of the sensitive feature.
This paper first studies one popular no disparate mistreatment fairness measure (i.e., OMR fairness) that requires the same misclassification rates for the different groups. We extend the results to two other no disparate mistreatment fairness measures--false positive rate and equal opportunity, and a popular no disparate impact fairness measure--demographic parity.

\noindent\textbf{Fair Classification:}
Fair classification can be achieved through pre-processing \citep{hajian2012methodology,calmon2017optimized,kamiran2012data}, in-processing \citep{kamishima2012fairness,agarwal2018reductions,zafar2017fairness}, or post-processing \citep{hardt2016equality,pleiss2017fairness,fish2016confidence} approaches. In this paper, we focus on the most flexible method (i.e., in-processing approach) to generate fairness-aware classifiers.
To reduce discrimination in machine learning, many recent studies incorporate fairness measures as constraints \citep{donini2018empirical,menon2018cost} or as a part of the objective function \citep{balashankar2019fair,aghaei2019learning}. For instance, fairness constraints of demographic parity \citep{menon2018cost,goel2018non,olfat2017spectral} or equal opportunity \citep{hardt2016equality,zafar2017fairness,menon2018cost} were added to the model formulations, while the fairness measures were suggested by \citep{balashankar2019fair} to be added into the objective function as a regularization term.
Due to nonconvexity of the most fairness measures, the fairness-related regularization or constraints usually were approximated to be convex ones \citep{olfat2017spectral,agarwal2018reductions}, which can be then reduced as efficiently solvable convex programs. 
These approximations are often inexact and can result in less fair results. 
Different from existing ones, this paper enforces the exact fairness by building mixed-integer programs, which enable to compute the fairness measures precisely. When incorporating fairness measures to construct fairness-aware classifiers, most of the existing works were only designated for a specific classifier. \Cref{literature_review} summarizes recent studies. As far as we are concerned, there is no existing approaches that are applicable to multiclass classification or black-box classifiers under fairness. The framework we proposed is flexible and can be extended to different types of fairness measures or different classifiers. It is also applicable to black-box classifiers. For example, it can be adapted to improve the fairness of popular deep classification models. Notably, our approach also shows advantages in the numerical study.   

\begin{table}[]
	\centering
	\setlength{\tabcolsep}{5pt}\renewcommand{\arraystretch}{1.2}
	\begin{tabular}{c|c|c|c|c|c}
		\hline
		\multirow{3}{*}{Existing Works} &
		\multirow{3}{*}{SVM} &
		\multirow{3}{*}{\begin{tabular}[c]{@{}c@{}}Kernel\\ SVM\\ \end{tabular}} &
		\multirow{3}{*}{\begin{tabular}[c]{@{}c@{}}Multiclass\\ SVM\\ \end{tabular}} &
		\multirow{3}{*}{\begin{tabular}[c]{@{}c@{}}Logistic\\ Regression\\ \end{tabular}} &
		\multirow{3}{*}{\begin{tabular}[c]{@{}l@{}}Black-box \\Classifiers\\ (e.g., CNN)\end{tabular}} \\
		&  &  &  &  &  \\ 
		&  &  &  &  &  \\ \hline	    
		\cite{donini2018empirical}, etc.     &$\checkmark$    &$\checkmark$    &  &     &  \\ \hline
		\begin{tabular}{c}
		\cite{zafar2019fairness},\\ \cite{olfat2017spectral},\\ \cite{zafar2017fairness}, etc.     
		\end{tabular}
		&$\checkmark$    &$\checkmark$    &  &$\checkmark$     &  \\ \hline
		\begin{tabular}{c}
		\cite{goel2018non},\\ \cite{kamishima2012fairness},\\ \cite{menon2018cost},\\ \cite{agarwal2018reductions},\\ \cite{taskesen2020distributionally}, etc.   
	\end{tabular}
		&  &  & &$\checkmark$   &  \\ \hline
		This paper &$\checkmark$    &$\checkmark$    &$\checkmark$   &$\checkmark$   &$\checkmark$    \\ \hline
	\end{tabular}
	\caption{Summary of Recent Studies on Fair Classification}
	\label{literature_review}
\end{table}
\vspace{10pt}
\subsection{Summary of Contributions}
\begin{figure}
	\centering
	\begin{subfigure}[b]{0.49\textwidth}
		\centering
		\includegraphics[width=1.0\linewidth]{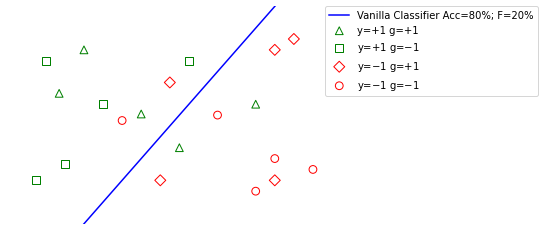}
		\caption{Decision Boundary for Vanilla Classifier}\label{demo1}
	\end{subfigure}
	\hfill
	\begin{subfigure}[b]{0.49\textwidth}
		\centering
		\includegraphics[width=1.0\linewidth]{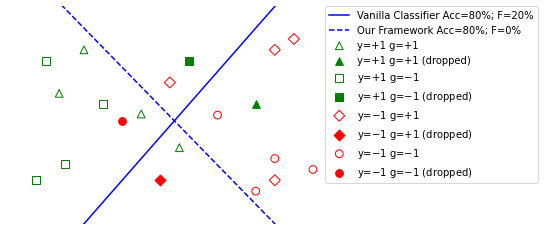}
		\caption{Decision Boundary for Our Framework}\label{demo2}
	\end{subfigure}
	\caption{The Role of Unbiased Subdata Selection: A Comparison of Vanilla Classifier and Our Framework. Note that $y\in \{-1,1\}$ denotes the class labels, $g\in \{-1,1\}$ denotes the protected groups, $Acc$ denotes the classification accuracy, and $F$ denotes the overall misclassification rate (OMR) (lower $F$ is better).}
	\label{fig:illustrate_subdata_selection}
	\vspace{-10pt}
\end{figure}

This paper first studies fair support vector machine (SVM) as a motivating example. We formulate the generalized SVM (GSVM) model as a mixed-integer program to model exact fairness, and then study the model properties, develop scalable algorithms for solving the generalized SVM with fairness (GSVMF) as well as study the extensions to other popular classification models including deep learning with different fairness measures. Specifically, our framework can be formulated as
\begin{equation} \label{generalized}
	\begin{aligned}
		\underset{(\bfw, b, \bfu) \in U,\bfz}{\min} &\left\{\underbrace{\A(\bfz, t\e-\bfu)}_{\text{Classification Outcome}}+ \lambda\underbrace{\R(\bfw,b,\bfu)}_{\text{Regularization}}+ \rho\underbrace{\F(\bfz)}_{\text{Fairness Measure}}:\bfz \in Z \right\},
	\end{aligned}
\end{equation}
where $t\in \Re$ is the prediction threshold, vector $\bfu$ represents classification outcomes (e.g., violation margins, prediction probabilities) under a particular classifier specified by unknown parameters {$\bfw, b$}, $\R(\bfw, b, \bfu)$ denotes the regularization term for the classifier with tuning parameter $\lambda\geq 0$, binary vector $\bfz \in Z$ indicates the correct classification corresponding to subdata selection decisions, and binary set $Z$ denotes the domain of $\bfz$. The binary vector $\bfz$ can be used to model the statistical fairness measures precisely without any approximation.
In the unified framework \eqref{generalized}, function $\A(\bfz, t\e-\bfu)$, which is often bilinear, denotes the classification outcomes and the function $\F(\bfz)$ denotes a fairness measure, where a Pareto optimality between accuracy and fairness is achieved by a proper penalty parameter $\rho>0$. \Cref{model_ex} displays the function $\A(\bfz, t\e-\bfu)$ and function $\R(\bfw, b, \bfu)$ for well-known classifiers. \Cref{fairness_ex} shows function $\F(\bfz)$ for four popular fairness measures. 
Besides, since the optimized binary decisions $\bfz$ take values $\{0,1\}$ according to the tradeoff of accuracy and fairness, the proposed method has the subdata selection interpretation. That is, we will select the data points such that their $z$ values are equal to one and train the classifier only using the selected subdata. Therefore, the proposed framework can select unbiased data points for fair classification and can be naturally solved using the iterative refining strategy with two subroutines, i.e., subdata selection and classification using the selected subdata. We illustrate the subdata selection in \Cref{fig:illustrate_subdata_selection}. Note that \Cref{demo1} demonstrates the decision boundary for the vanilla classifier which may not be fair to the protected groups, while \Cref{demo2} shows that the unbiased subdata selection in the proposed framework can reduce the unfairness score to 0\% while achieving the same accuracy as vanilla classifier.


\begin{table}[]
	\centering
	\setlength{\tabcolsep}{1.5pt}\renewcommand{\arraystretch}{1.5}
	\begin{tabular}{c|c|c}
		\hline Model & $\A(\bfz, t\e-\bfu) $ & $\R(\bfw,b,\bfu) $\\ \hline
		SVM  \eqref{GSVMF}  & $1/N\sum_{i\in[N]} z_i(u_i-t)$   & $\|\bfw\|^{2}_2$ \\ \hline
		Multiclass SVM \eqref{eq_GMSVM}   & $1/N\sum_{i\in[N]} \sum_{j\in[K],j \neq y_i} (1-z_{ij})(u_{ij}-t)$   & $\sum_{j\in[K]} \|\bfw_j\|_2^2$  \\ \hline
		
		%
		%
		Logistic Regression \eqref{GLRF} & $1/N\sum_{i\in[N]} z_i\left(t-u_i\right)$ & $\|\bfw\|^{2}_2$\\ \hline
		Deep Learning \eqref{cnn}  &$1/N\sum_{i\in[N]}  z_i\left(t-u_i\right)$    & 0  \\ \hline
	\end{tabular}
	\caption{Examples of Different Models under the Unified Framework \eqref{generalized}}
	\label{model_ex}
\end{table}

\begin{table}[]
	\vspace{-10pt}
	\centering
	\setlength{\tabcolsep}{5pt}\renewcommand{\arraystretch}{1.5}
	\begin{tabular}{c|c}
		\hline
		Fairness Measures & $\F(\bfz)$ \\ \hline
		Overall Misclassification Rate \eqref{Fair_function}  &$|\sum_{i\in \D_+}z_i/D_+ -\sum_{i\in \D_-}z_i/D_- |$   \\ \hline
		
		False Positive Rate \eqref{Fair_function_fpr}  & $| \sum_{i\in \D_{+-}}z_i/D_{+-}-\sum_{i\in \D_{--}}z_i/D_{--}|$  \\ \hline
		
		Equal Opportunity \eqref{Fair_function3}   & $|\sum_{i\in \D_{++}}z_{i}/D_{++}-\sum_{i\in \D_{-+}}z_{i}/D_{-+}|$  \\ \hline	
		\multirow{2}{*}{Demographic Parity \eqref{Fair_function2} } &$|\sum_{i\in \D_{++}}z_{i}/D_+ +\sum_{i\in \D_{+-}}(1-z_{i})/D_+$                       \\
		& \multicolumn{1}{l}{$-\sum_{i\in \D_{-+}}z_{i}/D_- -\sum_{i\in \D_{--}} (1-z_{i})/D_- |$} \\ \hline
	\end{tabular}
	\caption{Examples of Different Fairness Measures Expressed as $\F(\bfz)$}
	\label{fairness_ex}
	\vspace{-5pt}
\end{table}

%

\begin{figure}[h] 
	\centering
	\begin{tikzpicture}[ every text node part/.style={align=center}]
		
		\node[state,  fill=red!15] (th1) at (0.0,2.6) {A Unified Framework \eqref{generalized} to  \\ Incorporate Exact Fairness};
		\node[state, rectangle] (r1) at (-3.2, 1.1) {SVM};
		\node[state, rectangle] (r2) at (-3.2, -0.1) {\small{Multiclass} \\ \small{SVM}};
		\node[state, rectangle] (r3) at (-3.2, -1.3) {\small{Logistic} \\ \small{Regression}};
		\node[state, rectangle] (r4) at (-3.2, -2.5) {\small{Deep} \\ \small{Learning}};
		
		\node[state, rectangle, fill=cyan!15] (r0) at (0.0,-0.7) {\underline{Fairness Measures:} \\ \small{Overall Misclassification} \\ \small{Rate (\Cref{sec_GSVMF})} \\ \small{False Positive Rate }\\ \small{(\Cref{fpr_fair})}  \\ \small{Equal Opportunity} \\ \small{(\Cref{eo_fair})}\\ \small{Demographic Parity} \\ \small{(\Cref{dp_fair})}  };
		
		\draw[->,line width=0.7] (r1) -- (-2,1.1);
		\draw[->,line width=0.7] (r2) -- (-2,-0.1);
		\draw[->,line width=0.7] (r3) -- (-2,-1.3);
		\draw[->,line width=0.7] (r4) -- (-2,-2.5);

		\node[state, rectangle] (r5) at (3.5, 1.1) {GSVMF \\ \small{(\Cref{sec_GSVMF})}};
		\node[state, rectangle] (r6) at (3.5, -0.1) {GMSVMF \\ \small{(\Cref{sec_GMSVM})}};
		\node[state, rectangle] (r7) at (3.5, -1.3) {GLRF \\ \small{(\Cref{sec_lr})}};
		\node[state, rectangle] (r8) at (3.5, -2.5) {GDLF \\ \small{(\Cref{sec_faircnn})}};
		
		\draw[->,line width=0.7] (2,1.1) -- (r5);
		\draw[->,line width=0.7] (2,-0.1) -- (r6);
		\draw[->,line width=0.7] (2,-1.3) -- (r7);
		\draw[->,line width=0.7] (2,-2.5) -- (r8);
		
		\node[draw=red,fit=(r1) (r2) (r3) (r4) (r0) (r5) (r6) (r7) (r8), inner sep=0.1cm] (rel){};
		
		\draw [-, dashed] (4.85,-3.25) -- (4.85,3.55);
		
		\node[state,  fill=blue!10] (th2) at (8.15 , 2.6) {Efficient Algorithms};

		\node[state, rectangle, fill=orange!15] (a1) at (6.6, -1.3) {\footnotesize{Subdata Selection}\\ \footnotesize{Given Classification}\\ \footnotesize{Outcomes $\ast$}};
		\node[state, rectangle, fill=orange!15] (a2) at (9.7, -1.3) {\footnotesize{Optimize Loss} \\ \footnotesize{Using Selected} \\ \footnotesize{Data Points $\dagger$}};
		\draw[->,line width=0.7] (8.14, -1.1) -- (8.53, -1.1);
		\draw[->,line width=0.7] (8.53, -1.5) -- (8.15, -1.5);

		\node[state, rectangle] (o1) at (6.6, -0.1) {\small{\Cref{FE_GSVMF} $\ast$}};
		\node[state, rectangle] (o2) at (6.6, -2.5) {\small{\Cref{FE_GSVMF} $\ast$}};
		\node[state, rectangle] (o3) at (9.7, -0.1) {\small{\Cref{AM_alg} $\dagger$}};
		\node[state, rectangle] (o4) at (9.7, -2.5) {\small{\Cref{cnn_alg} $\dagger$}};
		\node[state, rectangle, fill=orange!15] (o0) at (8.15 , 1.1) {Iterative Refining Strategy\\ (\Cref{sec_AM})};

		\draw[->,line width=0.7] (7.8, 0) -- (8.49, 0);
		\draw[->,line width=0.7] (8.5, -0.2) -- (7.81, -0.2);
		
		\draw[->,line width=0.7] (7.8, -2.4) -- (8.49, -2.4);
		\draw[->,line width=0.7] (8.5, -2.6) -- (7.81, -2.6);
		
		%
		
		\draw[->,red,line width=1.1] (r8) -- (o2);
		
		\draw[decorate,red, decoration={brace, amplitude=10pt},line width=1.1] (4.5,1.6) -- coordinate [left=-10pt] (B) (4.5,-1.8) node {};
		\draw[->,red,line width=1.1] (B) -- (o1);

		\node[draw=blue,fit=(o1) (o2) (o3) (o4) (a1) (a2), inner sep=0.1cm] (rel){};
		\node[draw, fit= (th1) (th2) (r1) (r2) (r3) (r4) (r0) (r5) (r6) (r7) (r8) (o1) (o2) (o3) (o4) (a1) (a2), inner sep=0.2cm] (figure){};        
		
	\end{tikzpicture}
	\caption{The Roadmap of Our Contributions in This Paper. Note that \Cref{FE_GSVMF} is subdata selection procedure, and \Cref{AM_alg} and  \Cref{cnn_alg} employ the classifiers on the given the selected subdata.}\label{roadmap}
	\vspace{-1em}
\end{figure}
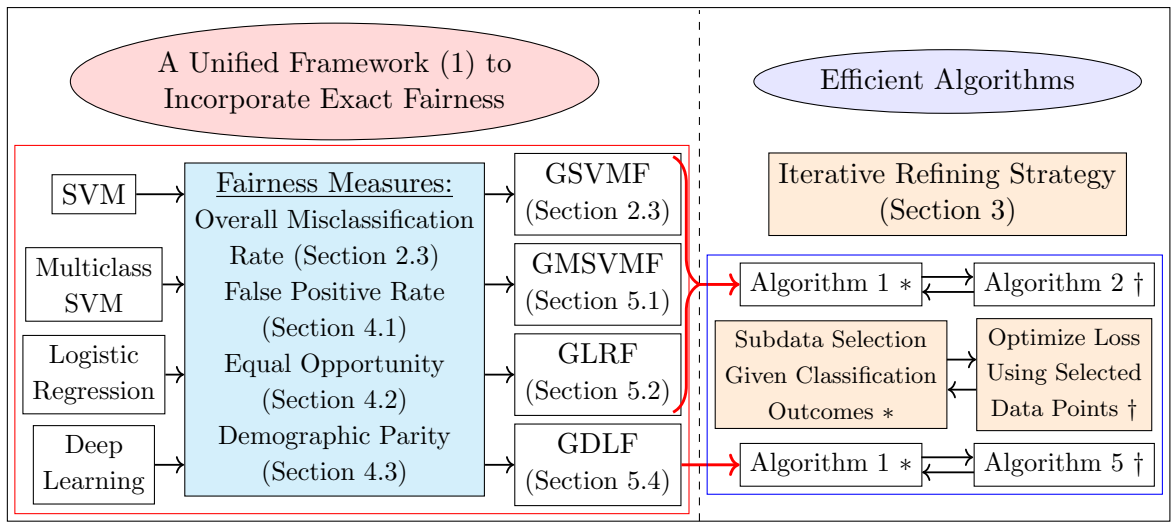

As illustrated in Figure  \ref{roadmap}, the main contributions of this paper are summarized below:
\begin{enumerate}[(i)]
	\item We propose a generalized SVM (GSVM) model for the binary classification problems and prove that it is Fisher consistent. We incorporate the overall misclassification rate (OMR) precisely  in the generalized SVM model, termed ``GSVMF"; 
	\item We propose an exact mixed-integer conic programming formulation for GSVMF, which can be solved  by off-the-shelf solvers for the moderate-sized instances; 
	\item For the large-scale instances, we propose to solve GSVMF effectively to near-optimality using the iterative refining strategy (IRS) by employing an unbiased subdata selection when fixing the classification outcomes and executing a classifier given the selected subdata in an alternative fashion. In the proposed IRS, the subdata selection, albeit resorting to a binary program, is strongly polynomial-solvable with time complexity of $O(N\log{N})$, where $N$ denotes the number of data points. We also study the convergence property of IRS and derive its approximation bound; 
	\item We show that the proposed GSVMF is also amenable to represent other fairness measures such as false positive rate, equal opportunity, and demographic parity; and 
	\item We further present the extensions to fair multiclass classification, logistic regression with fairness, kernel SVM with fairness, fair deep learning, and classification with unbalanced data. 
\end{enumerate}
The remainder of this paper is organized as follows. Section \ref{sec_model} presents the model formulation and model properties of GSVMF. Section \ref{sec_AM} studies the unbiased subdata selection method and the iterative refining strategy for solving GSVMF. Section \ref{sec_extension1} and Section \ref{sec_extension2} demonstrate the extensions to other fairness measures and different classifiers, respectively. Section \ref{sec_unbalanced} presents the application to classification with unbalanced data. Section \ref{sec_numerical} shows the numerical experiments. Section \ref{sec_conclusion} concludes the paper.

\noindent {\em Notation:} The following notation is used throughout the paper. We use bold-letters (e.g., $\bfx$, $\bfu$) to denote vectors or matrices, and use corresponding non-bold letters to denote their components. Given a real number $t$, we let $\lceil t\rceil$ be its round-up and $\lfloor t\rfloor$ be its round-down. Given an integer $n$, we let $[n]:=\{1,2,\ldots,n\}$. The indicator function $\I(\bm x \in R) =1$ if $\bm x \in R$, and $0$, otherwise. Given a vector $\bm{x}\in \Re^n$, we let $\supp (\bm{x})$ denote its support, i.e., $\supp (\bm{x}):=\{i:x_i\neq 0\}$. Given two subsets $S,T\subseteq [n]$, we let $S\Delta T$ denote the symmetric difference of sets $S$ and $T$, i.e., $S\Delta T:=(S\setminus T)\cup (T\setminus S)$.
Additional notation will be introduced as needed. 

\section{A Motivating Special Case: Generalized Support Vector Machine with OMR Fairness}\label{sec_model}
In this section, we will present model formulations and analyze properties for generalized support vector machine (GSVM) and extend GSVM with OMR fairness measure. Results of this section can be generalized to those having a different classifier or a distinct fairness measure.

Support vector machine (SVM) is a traditional approach for binary classification problems \citep{suykens1999least}. The aim of SVM is to construct a classifier to predict labels, i.e., the predictor $\tilde\bfx \in\mathbb{R}^n$ is used to predict the label $\tilde y\in\left\{-1,1\right\}$. Given a dataset $\{(\bfx_i,y_i)\}_{i\in [N]}\subseteq \Re^n\times \{-1,1\}$ with $N$ data points, SVM admits the following form
\begin{equation}\label{eq_svm}
	\underset{\bfw,b,\bfu}{\min} \left\{\sum_{i\in{[N]}}u_{i} + \lambda\|\bfw\|^{2}_2:
	y_i(\bfx^{\top}_i \bfw+b)\geq 1-u_i,
	u_i\geq 0,\forall{i\in{[N]}}\right\},
\end{equation}
where $\lambda\in\mathbb{R}$ is a tuning parameter, and $\bfw\in\mathbb{R}^n$, $b\in\mathbb{R}$ are unknown parameters of the classifier. 

\subsection{Generalized SVM (GSVM) and Mixed-Integer Programming Formulation}
Note that in SVM \eqref{eq_svm}, for each data point $i\in [N]$, the violation margin $u_i$ helps predict the classification results, i.e., if $u_i$ is less than or equal to some threshold, then we predict $i$th data point to be of the same label as $y_i$, and misclassify it, otherwise. Thus, this observation inspires us to improve SVM using an indicator function $\mathbbm{1}\left(u_i\leq t\right)$ to minimize the sum of misclassification margins, where t is the threshold. Specifically, we propose the following generalized SVM (GSVM)
\begin{equation}\label{eq_Gsvm_1}
	v:=\underset{\bfw, b ,\bfu}{\min} \left\{\frac{1}{N}\sum_{i\in[N]}(u_i-t)\mathbbm{1}\left(u_i\leq t\right)+\lambda\|\bfw\|^{2}_2:
	y_i(\bfx^{\top}_i \bfw+b)\geq 1-u_i,
	u_i\geq 0,\forall{i\in{[N]}} \right\},
\end{equation}
We remark that (i) if $t\rightarrow +\infty$, GSVM \eqref{eq_Gsvm_1} becomes SVM \eqref{eq_svm}; (ii) one can choose $t$ properly through cross-validation; and (iii) although being intractable, GSVM \eqref{eq_Gsvm_1} can be useful to incorporate many fairness measures and we will show later that it can be solved effectively via an off-the-shelf solver for moderate-sized instances and an iterative refining strategy for large-scale instances.

Next, we observe that in GSVM \eqref{eq_Gsvm_1}, for each $i\in [N]$, the indicator function $\mathbbm{1}\left(u_i\leq t\right)$ can be replaced by the binary variable $z_i$, where $z_i=1$ indicates correct classification and $z_i=0$ indicates misclassification with respect to the threshold $t$. We see that, GSVM \eqref{eq_Gsvm_1} admits an equivalent mixed-integer biconvex programming formulation.
\begin{proposition}
	GSVM is equivalent to
	\begin{subequations} \label{eq_biconvex}\label{GSVM}
		\begin{align}
			v:=\underset{\bfw,b,\bfu,\bfz}{\min} &\frac{1}{N}\sum_{i\in[N]} z_i(u_i-t)+\lambda\|\bfw\|^{2}_2,\\
			\text{\rm(GSVM)} \qquad
			\text{s.t.} \quad &y_i(\bfx^{\top}_i \bfw+b)\geq 1-u_i,\forall{i\in{[N]}} \label{eq_plane},\\ 
			&u_i\geq 0,\forall{i\in{[N]}},\label{eq_u}\\
			&z_i\in \{0,1\},\forall{i\in{[N]}} \label{eq_binary_z},
		\end{align}
	\end{subequations}

\end{proposition}
\proof The proof follows by observing that at optimality,  for each $i\in [N]$, we must have $z_i=1$ if $u_i\leq t$, and 0, otherwise; that is, $z_i=\mathbbm{1}\left(u_i\leq t\right)$. \QEDA

The binary variables $\bfz$ enable us to distinguish the correct classified and misclassified data points. Thus, we can model the fairness using these binary variables. 
Note that the biconvex program \eqref{eq_biconvex} can be equivalently reformulated as a mixed-integer conic program (MICP) with the following McCormick inequalities to linearize the bilinear terms $\{z_iu_i\}_{i\in [N]}$
\begin{equation}\label{MIP_constr1} 
	s_i\leq M_u z_i, \quad s_i\leq u_i, \quad s_i\geq 0, \quad s_i\geq u_i-M_u(1-z_i), \forall i\in [N]
\end{equation}
where $M_u$ is a positive large number and its derivation will be postponed to the next proposition. The result of MICP is summarized below.
\begin{proposition} \label{prop_micp_gsvm}
	GSVM can be formulated as the following MICP
	\begin{equation}\label{eq_micp} 
		v:=\underset{\bfw, b ,\bfu,\bfz,\bm{s}}{\min} \left\{\frac{1}{N}\sum_{i\in[N]}(s_i-z_i t)+\lambda\|\bfw\|^{2}_2:
		\eqref{eq_plane}-\eqref{eq_binary_z},\eqref{MIP_constr1}\right\}.
	\end{equation}
\end{proposition}
This result allows us to use off-the-shelf solvers such as Gurobi and CPLEX to solve \eqref{eq_micp} to optimality when the number of data points is not very large, and we can also use the optimal solution from this exact formulation to test the effectiveness of the proposed iterative refining strategy.

The next proposition shows how to compute the big-M coefficient $M_u$ from GSVM \eqref{eq_micp} in closed-form. The proof idea is to (i) use the optimality condition; and (ii) discuss different cases of label $y_i$ for each $i\in [N]$.
\begin{proposition}\label{big_M}
	Suppose that $L_1=\sqrt{t\lambda^{-1}}$ and $L_2=\max_{i\in [N]}\|\bfx_i\|_2$. Then $M_u$ can be chosen as $M_u=2+2L_1L_2$.
\end{proposition}
\proof 
First of all, observe that the optimal value of GSVM \eqref{eq_biconvex} must satisfy $v\leq 0$ since $\bfw=\bm0$, $b=0$, $\bfu=t$, $\bfz=0$ is a feasible solution with objective value equal to 0. Thus, there exists an optimal solution $(\bfw^*,b^*,\bfu^*,\bfz^*)$ such that 
$$\frac{1}{N} \sum_{i\in[N]} z^*_i(u^*_i-t)+\lambda\|\bfw^*\|^{2}_2 \leq 0.$$ Since $ \sum_{i\in[N]} z_i^*(u_i^*-t) \geq -Nt$, thus $\lambda\|\bfw^*\|^{2}_2 \leq t$. Therefore, we can upper bound $\|\bfw^*\|_2$ as  $\|\bfw^*\|_2\leq \sqrt{t\lambda^{-1}}=L_1$. 
Next, we are going to derive the bound $M_b$ for $|b^*|$. 
Now letting $a_i:=-y_{i} (\bfx_{i}^{\top}\bfw+b)+1-t$, we have $u_i=\max\{a_i+t,0\}$ for each $i\in [N]$. Projecting out variables $\bm u$ and using the fact that $\max\{a_i+t,0\}\leq t$ is equivalent to $a_i\leq 0$ since $t\geq 0$, GSVM \eqref{eq_Gsvm_1} is equivalent to
$$v :=\underset{\bfw,b}{\min} \left\{\frac{1}{N}\sum_{i\in[N]}\max\{a_i:=-y_{i} (\bfx_{i}^{\top}\bfw+b)+1-t,-t\}\mathbbm{1}(a_i\leq0)+\lambda\|\bfw\|^{2}_2 \right\}.$$
Let us denote $a_i^*=-y_i(\bfx_i^{\top} \bfw^*+b^*)+1-t$. We find $M_b$ by discussing the following four cases for each $i\in [N]$:
\begin{enumerate}[{Case} i.]
	\item  If $y_i=+1$ and $a_i^*\leq-t$, then we must have $y_i(\bfx^{\top}_i \bfw^*+b^*)\geq1$. According to Cauchy-Schwartz inequality $|\bfx_{i}^{\top}\bfw|\leq L_1L_2$ and triangle inequality, this inequality (i.e., $y_i(\bfx^{\top}_i \bfw^*+b^*)\geq1$) will be guaranteed by choosing $M_b\geq1+L_1L_2$.
	\item  If $y_i=+1$ and $a_i^*\geq-t$, then we must have $y_i(\bfx^{\top}_i \bfw^*+b^*)\leq1$. It is sufficient to choose $M_b \geq 1+L_1 L_2$ to guarantee the inequality.
	\item  If $y_i=-1$ and $a_i^*\leq-t$, then we must have $y_i(\bfx^{\top}_i \bfw^*+b^*)\geq1$. It is sufficient to choose $M_b\geq 1+L_1 L_2$ to guarantee the inequality.
	\item  If $y_i=-1$ and $a_i^*\geq-t$, then we must have $y_i(\bfx^{\top}_i \bfw^*+b^*)\leq1$. It is sufficient to choose $M_b\geq1+L_1 L_2$ to guarantee the inequality.
\end{enumerate}
Finally, to ensure that the constraint \eqref{eq_plane} holds for each $i\in [N]$, according to Cauchy-Schwartz inequality $|\bfx_{i}^{\top}\bfw|\leq L_1L_2$, it is sufficient to choose 
$$M_u \geq |1-y_i(\bfx^{\top}_i \bfw+b)|\quad\Longleftarrow \quad M_u \geq 1+M_b+L_1 L_2\geq 1+|b^*|+L_1 L_2.$$ Therefore, $M_u = 1+M_b+L_1 L_2=2+2L_1L_2$ suffices. \QEDA 

\subsection{Fisher Consistency of GSVM}
In this subsection, we prove Fisher consistency of the proposed GSVM \eqref{GSVM}, which further demonstrates the correctness of the proposed model. Given a realization $\bm{x}$ of random observation $\tilde{\bm{x}}$, let us define
$$p(\bm{x})=\Pr\left\{\tilde{y}=1\mid \tilde{\bm{x}}=\bm{x}\right\},$$
i.e., $p(\bm{x})$ defines the probability that the label of data $\bm{x}$ equals to $1$. Following \citet{lee2004multicategory,lin2002support,liu2007fisher}, we consider the Bayesian decision rule  $\phi_B(\bm{x})$, which admits the following form
\begin{align}
	\phi_B(\bm{x})=\begin{cases}
		1, \text{\rm if } p(\bm{x})\geq 0.5,\\
		-1, \text{\rm if } p(\bm{x})<0.5,
	\end{cases}
\end{align}
i.e., we assign a label  for the realization $\bm{x}$ according to whether the probability $p(\bm{x})$ is greater than 0.5 or not. In formulation \eqref{eq_biconvex}, we will use a more general classifier $f(\bm{x})$ instead of the linear one $\bm{x}^\top \bm{w}+b$. Conditioning on the realization $\tilde{\bm x}=\bm{x}$ in formulation \eqref{GSVM}, we are trying to find a separator $f(\bm{x})$ to minimize the conditional expected surrogate loss. Given a realization $\bm{x}$ of the random data, we suppose that the classifier $f(\bm{x})$ is normalized to be between $-t-1$ and $t+1$. The classification problem is equivalent to
\begin{align}
	\min_{f(\bfx), \tilde{u}} \left\{\E[(\tilde{u}-t)\mathbbm{1}\left( \tilde{u}\leq t\right)|\tilde{\bfx}=\bfx]:\tilde{y}f(\tilde{\bfx})\geq 1-\tilde{u}, \tilde{u}\geq 0, -t-1\leq f(\bm{x})\leq t+1\right\}.\label{eq_loss0}
\end{align}
Projecting out random variable $\tilde{u}$, formulation \eqref{eq_loss0} is equivalent to 
\begin{equation}
\begin{aligned}
	\min_{f(\bfx)}& \left\{ L(f(\bm{x})):=\E_{\tilde{y}}\left[\max\{-\tilde{y} f(\tilde{\bfx})+1-t,-t\}\mathbbm{1}(\tilde{y} f(\tilde{\bfx})\geq t-1)|\tilde{\bm x}=\bm{x}\right]:\right.\\
	&\left.-t-1\leq f(\bm{x})\leq t+1 \right\}.\label{eq_loss}
\end{aligned}
\end{equation}
Thus, the expected loss minimization problem can be formally defined as
\begin{align}
	f^*(\bm{x}) \in \arg\min_{f(\bm{x})}\left\{L(f(\bm x)): -t-1\leq f(\bm{x})\leq t+1\right\}, \label{eq_f_L}
\end{align}
where we let $f^*(\bm{x})$ denote an optimal classifier.\par

Next, following the spirit of seminal works \citep{lee2004multicategory,lin2002support,liu2007fisher}, we make a formal definition of Fisher consistency for the binary classification problem.
\begin{definition}
	(\textbf{Fisher Consistency}) For any realization $\bm{x}$ of random data $\tilde{\bm{x}}$, its loss function $L(\bm{f}(\bm{x}))$ is \textit{Fisher consistent} if
	\[\phi_B(\bm{x})=2\I(p(\bm{x})\geq 0.5)-1=2\I(\bm{f}^*(\bm{x})\geq t-1)-1,\]
	where $\bm{f}^*(\bm{x})$ is defined in \eqref{eq_f_L}.
\end{definition}

Now we are ready to show the consistency of the loss function $L(\bm{f}(\bm{x}))$.
\begin{proposition}\label{prop_bayes_consis} Given $t>0$, the loss function $L(\bm{f}(\bm{x}))$ defined in \eqref{eq_loss} is Fisher consistent.
\end{proposition}
\proof
Note that according to \eqref{eq_loss}, we have
\begin{align*}
	L(f(\bm{x}))&=p(\bfx)\max\{-f({\bfx})+1-t,-t\}\mathbbm{1}( f(\bfx)\geq t-1)\\
	&+(1-p(\bfx))\max\{f({\bfx})+1-t,-t\}\mathbbm{1}( f(\bfx)\leq 1-t).
\end{align*}
Thus, minimizing the loss function in \eqref{eq_f_L} is equivalent to solving the following minimization problem
\begin{align*}
	&\min_{f(\bm{x})}\left\{L(f(\bm{x})):-t-1\leq f(\bfx)\leq t+1\right\}
\end{align*}
where a minimizer is $f^*(\bm{x})=t+1$ if $p(\bm{x})\geq 0.5$, and $-t-1$, otherwise. Thus, we have
\[\phi_B(\bm{x})=2\I(p(\bm{x})\geq 0.5)-1=2\I(\bm{f}^*(\bm{x})\geq t-1)-1.\]
\QEDA

\Cref{prop_bayes_consis} shows that similar to SVM \citep{lin2002support}, GSVM \eqref{GSVM} is also Fisher consistent. This demonstrates that the performance of the proposed GSVM \eqref{GSVM} is at least as good as SVM for the binary classification problems. We remark that the similar Fisher consistency result holds for the multi-class SVM.

\subsection{GSVM with Fairness (GSVMF)} \label{sec_GSVMF}
In this subsection, we will illustrate the modeling power of GSVM \eqref{GSVM} by augmenting it with one popular fairness measure, overall misclassification rate (OMR). Extensions to other fairness measures can be found in \Cref{sec_extension1}. 

The OMR (see, e.g., \citealt{zafar2019fairness}) fairness accounts for the misclassification measure of disparate mistreatment. The classifier has no disparate mistreatment if different groups have the same misclassification rates. Suppose that there is a protected feature corresponding to two groups denoted by $g\in \{-1,1\}$. Then OMR fairness measure can be defined as below.
\begin{definition} \label{df_omr}Given a threshold $t>0$, the OMR fairness measure for GSVM \eqref{eq_Gsvm_1} is
	$$F_{\rm OMR} = \left|\Pr(\tilde{u}\leq t|\tilde g=+1)-\Pr(\tilde{u}\leq t|\tilde g=-1)\right|.$$
\end{definition} 
The fairness measure function $F_{\rm OMR}$ is the absolute difference of the misclassification rates of both groups. We then use the binary variables $\bfz$ to exactly represent the OMR fairness measure
\begin{subequations}
	\begin{equation} \label{Fair_function}
		\begin{aligned}
			F_{\rm OMR} &= \left|\Pr(\tilde{u}\leq t|\tilde g=+1)-\Pr(\tilde{u}\leq t|\tilde g=-1)\right|
			= \left|\sum_{i\in \D_+}\frac{z_i}{D_+}-\sum_{i\in \D_-}\frac{z_i}{D_-}\right|,
		\end{aligned}
	\end{equation}
	where two sets
	\begin{equation}
		\D_+=\{i\in[N]:   g_i=+1\},\D_-=\{i\in[N]:   g_i=-1\}
	\end{equation}
\end{subequations}
represent index subsets of data points from each protected group, respectively, and $D_+=|\D_+|$ is the number of group $g=+1$ data points, and $D_-=|\D_-|$ is the number of group $g=-1$ data points. 

To prevent discrimination in prediction, we penalize the unfairness in our proposed GSVM \eqref{eq_biconvex} with nonnegative penalty $\rho$. Therefore, we formulate the following GSVMF model
\begin{equation}\label{GSVMF} 
	v_\rho:=\underset{\bfw, b ,\bfu,\bfz}{\min} \left\{ \underbrace{\frac{1}{N}\sum_{i\in[N]} z_i(u_i-t)}_{\A(\bfz, t\e-\bfu)}+\lambda  \underbrace{\|\bfw\|^{2}_2}_{\R(\bfw, b, \bfu)}+\rho\underbrace{\left|\sum_{i\in \D_+}\frac{z_i}{D_+}-\sum_{i\in \D_-}\frac{z_i}{D_-}\right|}_{\F(\bfz)}:\eqref{eq_plane}-\eqref{eq_binary_z}\right\}
\end{equation}
where $\rho$ is the penalty factor that balances the prediction accuracy and fairness. Clearly, the proposed GSVMF \eqref{GSVMF} is a special case of the framework \eqref{generalized}. 
Note that GSVMF \eqref{GSVMF} has non-convex terms $\{z_i u_i\}_{i\in [N]}$. We can linearize them using McCormick inequalities \eqref{MIP_constr1}. Thus, similar to \Cref{prop_micp_gsvm}, GSVMF \eqref{GSVMF} can be formulated as a MICP.
\begin{proposition} \label{cro_prop2}
	Formulation GSVMF is equivalent to
	\begin{align*}
		v_{\rho}:=\underset{\bfw, b ,\bfu,\bfz,\bm{s}}{\min} \left\{\frac{1}{N}\sum_{i\in[N]}(s_i-z_i t)+\lambda\|\bfw\|^{2}_2+\rho\left|\sum_{i\in \D_+}\frac{z_i}{D_+}-\sum_{i\in \D_-}\frac{z_i}{D_-}\right|:
		\eqref{eq_plane}-\eqref{eq_binary_z},\eqref{MIP_constr1}\right\}.
	\end{align*}
\end{proposition}

We remark that for the moderate-sized instances, one might be able to solve both MICPs easily via off-the-shelf solvers such as Gurobi and CPLEX.


The observation that GSVMF \eqref{GSVMF} is a special case of the framework \eqref{generalized} leads to the following remarks.

\begin{enumerate}[(i)]
	\item The binary variables $\bm{z}$ indicate the subdata selection decisions. Data points with their corresponding $z-$variables taking value of 1 are the potential unbiased ones selected by the model, and the other data points are ignored. This gives rise to an iterative refining strategy described in \Cref{sec_AM} that improves the subdata selection decisions $\bfz$ and classification accuracy in an alternating way; 
	\item The proposed GSVMF \eqref{GSVMF} can be quite flexible. Indeed, if we replace $u_i$ 
	by the probabilities of classification outcomes and let $t$ denote the misclassification threshold, then GSVMF \eqref{GSVMF} can be generalized to other classification models with fairness including logistic regression and convolutional neural networks (CNNs). These facts will be elaborated in the subsequent sections; 
	\item Since the classification results are discrete (e.g., $\{0,1\}$ for the binary classification), it is natural to model the fairness measures using discrete variables, for which GSVMF \eqref{GSVMF} paves a generic way; 
	\item For the moderate-sized datasets, we can use exact approaches to solve GSVMF \eqref{GSVMF} to optimality. For the large-scale optimization, we propose to use the iterative refining strategy (see \Cref{sec_AM}), which works extremely well in our numerical study;
	\item If one would like to add the OMR fairness measure as a constraint, the similar results still hold; and
	\item The OMR fairness measure in GSVMF \eqref{GSVMF} can be replaced by other fairness measures such as false positive rate (FPR), equal opportunity, and demographic parity, which will be elaborated in \Cref{sec_extension1}.
\end{enumerate}

We observe that the absolute term in the objective function in GSVMF \eqref{GSVMF} can be split into two parts by discussing its sign, and therefore GSVMF \eqref{GSVMF} can be computed by solving two distinct optimization problems. This observation will be particularly important when we design the iterative refining strategy for solving GSVMF \eqref{GSVMF} in \Cref{sec_AM}.
\begin{proposition} \label{prop2}
	The optimal value of GSVMF \eqref{GSVMF} is $v_{\rho}=\min\{v_{\rho1},v_{\rho2}\}$, where $v_{\rho1}$ and $v_{\rho2}$ are optimal values of the following two optimization problems
\end{proposition}
\begin{subequations}
	\begin{equation}\label{GSVMF_1-constr}
		\begin{aligned}
			v_{\rho1}:=\underset{\bfw, b ,\bfu,\bfz}{\text{min}} &\frac{1}{N}\sum_{i\in[N]} z_i(u_i-t)+\lambda\|\bfw\|^{2}_2-\rho\sum_{i\in \D_+}\frac{z_i}{D_+}+\rho\sum_{i\in \D_-}\frac{z_i}{D_-},\\
			\text{\rm (GSVMF-1)}\quad \text{s.t.}&\sum_{i\in \D_+}D_-z_i-\sum_{i\in \D_-}D_+z_i\leq0,\\
			&\eqref{eq_plane}-\eqref{eq_binary_z};\
		\end{aligned}
	\end{equation}
	\begin{equation}\label{GSVMF-2_constr}
		\begin{aligned}
			v_{\rho2}:=\underset{\bfw, b ,\bfu,\bfz}{\text{min}} &\frac{1}{N}\sum_{i\in[N]} z_i(u_i-t)+\lambda\|\bfw\|^{2}_2+\rho\sum_{i\in \D_+}\frac{z_i}{D_+}-\rho\sum_{i\in \D_-}\frac{z_i}{D_-},\\
			\text{\rm (GSVMF-2)}\quad \text{s.t.}&\sum_{i\in \D_+}D_-z_i-\sum_{i\in \D_-}D_+z_i\geq0,\\
			&\eqref{eq_plane}-\eqref{eq_binary_z}.
		\end{aligned}
	\end{equation}
\end{subequations}



\subsection{Connection to Zero-tolerance GSVMF}
In this subsection, 
we observe that if $\rho$ is large enough or more precisely, if there exists a threshold $\bar{\rho}$ such that $\rho\geq\bar{\rho}$, then the fairness measure becomes zero and GSVMF \eqref{GSVMF} becomes the following GSVMF0
\begin{equation} 
	\underset{\bfw, b ,\bfu,\bfz}{\min} \left\{\frac{1}{N} \sum_{i\in[N]} z_i(u_i-t)+\lambda\|\bfw\|^{2}_2: \left|\sum_{i\in \D_+}\frac{z_i}{D_+}-\sum_{i\in \D_-}\frac{z_i}{D_-}\right|=0, \eqref{eq_plane}-\eqref{eq_binary_z} \right\},\label{eq_GSVMF0}
\end{equation}
i.e., zero-tolerance GSVMF \eqref{GSVMF}. This result is summarized below.
\begin{proposition}\label{prop_gsvmf_barrho}
	There exists a positive $\bar{\rho}>0$ such that for any $\rho \geq \bar{\rho}$, GSVMF reduces to GSVMF0 \eqref{eq_GSVMF0}. In particular, $\bar{\rho}$ can be chosen as $\bar{\rho}=\frac{\hat{r}}{\hat{s}}$, where $\hat{r}>0$ and $\hat{s}>0$ are defined as below:
	\begin{align*}
		\hat{s}:&=\min_{\bm{z}\in \{0,1\}^N}\left\{\left|\sum_{i\in \D_+}\frac{z_i}{D_+}-\sum_{i\in \D_-}\frac{z_i}{D_-}\right|>0\right\},\\
		\hat{r}:&=\max_{\bm{z},\hat{\bfz}\in \{0,1\}^N}\left\{\min_{\bm{w},b,\bm{u}}\left\{\frac{1}{N} \sum_{i\in[N]}z_i(u_i-t)+\lambda\|\bfw\|^{2}_2:\eqref{eq_plane}-\eqref{eq_binary_z}\right\}\right.\\
		&\left.-\min_{\bm{w},b,\bm{u}}\left\{\frac{1}{N}\sum_{i\in[N]} \hat{z}_i(u_i-t)+\lambda\|\bfw\|^{2}_2:\eqref{eq_plane}-\eqref{eq_binary_z}\right\}> 0\right\}.
	\end{align*}
\end{proposition}


\begin{subequations}
	\proof 
	For any $\rho>0$, let $(\hat{\bm{w}},\hat{b},\hat{\bm{u}},\hat{\bm{z}})$ and $(\bm{w}^*,b^*,\bm{u}^*,\bm{z}^*)$ denote the optimal solutions of GSVMF \eqref{GSVMF} and GSVMF0 \eqref{eq_GSVMF0}, respectively; and let $\hat{v}_{\rho}$, $v^*$ denote their corresponding optimal objective values. To prove the result, it is equivalent to show that $\hat{v}_{\rho}=v^*$ for any $\rho \geq \bar{\rho}$, i.e., $(\bm{w}^*,b^*,\bm{u}^*,\bm{z}^*)$ is optimal to GSVMF \eqref{GSVMF}.
	
	Since $(\bm{w}^*,b^*,\bm{u}^*,\bm{z}^*)$ is feasible to GSVMF \eqref{GSVMF}, thus we have $v^*\geq \hat{v}_{\rho}$.
	Thus, it remains to show that $v^*\leq \hat{v}_{\rho}$ when $\rho\geq \bar{\rho}$, $\bar{\rho}>0$. We prove it by contradiction. Suppose $v^*> \hat{v}_{\rho}$, which implies that $(\bm{w}^*,b^*,\bm{u}^*,\bm{z}^*)$ is not optimal to GSVMF \eqref{GSVMF}, or equivalently, $(\hat{\bm{w}},\hat{b},\hat{\bm{u}},\hat{\bm{z}})$ is not optimal to GSVMF0 \eqref{eq_GSVMF0}. Hence, due to optimality condition, we must have
	\begin{align}
		&\left|\sum_{i\in \D_+}\frac{\hat{z}_i}{D_+}-\sum_{i\in \D_-}\frac{\hat{z}_i}{D_-}\right|>0.\label{eq_subopt}
	\end{align}
	Additionally, 
	\begin{align}\label{eq_subopt2}
		&\frac{1}{N} \sum_{i\in[N]} \hat{z}_i(\hat{u}_i-t)+\lambda\|\hat{\bfw}\|^{2}_2=\hat{v}_{\rho}-\rho\left|\sum_{i\in \D_+}\frac{\hat{z}_i}{D_+}-\sum_{i\in \D_-}\frac{\hat{z}_i}{D_-}\right|\\
		&<v^*-\rho\left|\sum_{i\in \D_+}\frac{\hat{z}_i}{D_+}-\sum_{i\in \D_-}\frac{\hat{z}_i}{D_-}\right| \notag <v^*:=\frac{1}{N} \sum_{i\in[N]} z^*_i(u^*_i-t)+\lambda\|\bfw^*\|^{2}_2,
	\end{align}
	where the first inequality is due to our assumption and the second one is because of \eqref{eq_subopt}.
	
	Then according to \eqref{eq_subopt} and \eqref{eq_subopt2} as well as the definition of $\hat{s}$ and $\hat{r}$, we must have $\hat{s}$, $\hat{r}>0$. Hence, we can further conclude that
	\begin{equation}
		\hat{s}\leq \left|\sum_{i\in \D_+}\frac{\hat{z}_i}{D_+}-\sum_{i\in \D_-}\frac{\hat{z}_i}{D_-}\right|,
		\label{s*}
	\end{equation}  
	\begin{equation}
		\begin{aligned}
			\frac{1}{N} \sum_{i\in[N]}\hat{z}_i(\hat{u}_i-t)+\lambda\|\hat{\bfw}\|^{2}_2 &\geq \min_{\bm{w},b,\bm{u}}\left\{\frac{1}{N} \sum_{i\in[N]} \hat{z}_i(u_i-t)+\lambda\|\bfw\|^{2}_2:\eqref{eq_plane}-\eqref{eq_binary_z} \right\}\\
			&\geq \frac{1}{N} \sum_{i\in[N]}z^*_i(u^*_i-t)+\lambda\|\bfw^*\|^{2}_2-{\hat{r}}.
		\end{aligned}
		\label{v*}
	\end{equation}
	
	According to \eqref{s*} and \eqref{v*}, we obtain
	\begin{align*}
		\hat{v}_{\rho}&=\frac{1}{N} \sum_{i\in[N]} \hat{z}_i(\hat{u}_i-t)+\lambda\|\hat{\bfw}\|^{2}_2+\rho\left|\sum_{i\in \D_+}\frac{\hat{z}_i}{D_+}-\sum_{i\in \D_-}\frac{\hat{z}_i}{D_-}\right|\\
		&\geq\frac{1}{N} \sum_{i\in[N]}z^*_i(u^*_i-t)+\lambda\|\bfw^*\|^{2}_2-{\hat{r}}+\rho\hat{s}.
	\end{align*}
	Thus, for $\rho\geq\bar{\rho}:=\frac{{\hat{r}}}{\hat{s}}$, we must have $\hat{v}_{\rho} \geq v^*$, a contradiction.\QEDA
\end{subequations}

We remark that (i) the proof of \Cref{prop_gsvmf_barrho} highly relies on the discrete nature of variables $\bm{z}$, particularly, the reason that two constants $\hat{r}>0$ and $\hat{s}>0$ are well defined is because binary variables $\bm{z}$ can only take a finite number of values; and (ii) the result in \Cref{prop_gsvmf_barrho} shows that by choosing a large but finite penalty $\rho$ instead of being infinite, we are able to convert GSVMF \eqref{GSVMF} into fairness constrained GSVMF formulation.


\section{Iterative Refining Strategy (IRS): Theory and Implementations}\label{sec_AM}
For the large-scale datasets, exact methods might not be able to solve GSVMF \eqref{GSVMF} effectively, thus we propose to solve GSVMF \eqref{GSVMF} using iterative refining strategy (IRS), motivated by  alternating minimization method from optimization community. 
In the IRS, we will optimize binary variables $\bfz$ by fixing the values of the continuous variables $(\bfw,b,\bfu)$ in GSVMF \eqref{GSVMF}, termed ``unbiased subdata selection," fix the values of the binary variables $\bfz$ and optimize the continuous variables $(\bfw,b,\bfu)$; then iterate. In this section, we will study the unbiased subdata selection, design the IRS, and analyze the convergent property of the IRS.

\subsection{Unbiased Subdata Selection}\label{sec_ss}
In this subsection, we will develop a polynomial-time \Cref{FE_GSVMF} with running time complexity $O(N\log N)$ for solving GSVMF \eqref{GSVMF} by fixing the values of the continuous variables $(\bfw,b,\bfu)$. The algorithm takes advantage of the notable result in \Cref{prop2}, that is, we can decompose the problem into two subproblems, GSVMF-1 \eqref{GSVMF_1-constr} and GSVMF-2 \eqref{GSVMF-2_constr}. For each subproblem, we observe that if the total number of the correct classification outcomes of the group $g=+1$ is known, i.e., $\sum_{i\in \D_+}z_i=\textrm{\rm constant}$, then the resulting binary optimization problem can be solved effectively by sorting the objective coefficients. The formal derivation can be found below.
\begin{proposition}\label{prop_opt_z}
	For any fixed $(\bfw,b,\bfu)$ in GSVMF \eqref{GSVMF}, optimizing over $\bfz$ can be done in time complexity $O(N\log{N})$.
\end{proposition}
\proof Due to symmetry, it is sufficient to show that the restricted \text{\rm GSVMF-1} \eqref{GSVMF_1-constr} can be solved in $O(N\log{N})$ when the values of $(\bfw,b,\bfu)$ are fixed. We split the proof into three steps.

\noindent \textbf{Step 1. }Observe that to solve the restricted \text{\rm GSVMF-1} \eqref{GSVMF_1-constr}, we can first enumerate the value of $\sum_{i\in \D_-}z_i$, i.e., we can first let $\sum_{i\in \D_-}z_i=k_-$ for any integer $k_-\in \{0,1,\ldots, D_-\}$. Then the restricted \text{\rm GSVMF-1} \eqref{GSVMF_1-constr} is equivalent to solve the following two optimization problems
\begin{subequations}
	\begin{equation}\label{GSVMF_1-constr_sub1}
		\begin{aligned}
			\underset{\{z_i\}_{i\in \D_+}}{\text{min}} \left\{\sum_{i\in \D_+}z_i\left[(u_i-t)/N-\rho/D_+\right]:\sum_{i\in \D_+}z_i\leq \lfloor k_-D_+/D_-\rfloor\right\},
		\end{aligned}
	\end{equation}
	\begin{equation}\label{GSVMF_1-constr_sub2}
		\begin{aligned}
			\underset{\{z_i\}_{i\in \D_-}}{\text{min}} \left\{\sum_{i\in \D_-}z_i\left[(u_i-t)/N+\rho/D_-\right]:\sum_{i\in \D_-}z_i= k_-\right\},
		\end{aligned}
	\end{equation}
\end{subequations}
where both problems can be solved efficiently via sorting two lists $\{\hat{u}_i:=(u_i-t)/N\}_{i\in \D_+}$ and $\{\hat{u}_i:=(u_i-t)/N\}_{i\in \D_-}$ in the ascending order. Namely, suppose that $\{\hat{u}_i\}_{i\in \D_+}$ and $\{\hat{u}_i\}_{i\in \D_-}$ are sorted as $\hat{u}_{\sigma_+(1)}\leq...\leq \hat{u}_{\sigma_+(D_+)}$, $\hat{u}_{\sigma_-(1)}\leq...\leq \hat{u}_{\sigma_-(D_-)}$, respectively.

Now let us define $\tau_1^*=\arg\max_{i\in \D_+}\{\hat{u}_i-\rho/D_+<0\}$ and $k_+=\min\{\lfloor k_-D_+/D_-\rfloor,\tau_1^*\}$. Let
$z_i^*=1$ if $i\in \{\sigma_+(\ell)\}_{\ell\in [k_+^*]}\cup \{\sigma_-(\ell)\}_{\ell\in [k_-^*]}$ and 0, otherwise. Then $\{z_i^*\}_{i\in \D_+},\{z_i^*\}_{i\in \D_-}$ solve the problems \eqref{GSVMF_1-constr_sub1} and \eqref{GSVMF_1-constr_sub2}, respectively. 

\noindent \textbf{Step 2. } We choose the best $\bfz^*$ which achieves the smallest objective value among the possible ones for all $k_-\in \{0,1,\ldots, D_-\}$.

\noindent \textbf{Step 3. }Note that the sorting of two lists $\{\hat{u}_i:=(u_i-t)/N\}_{i\in \D_+}$ and $\{\hat{u}_i:=(u_i-t)/N\}_{i\in \D_-}$, and the calculations of $\{\hat{U}_{\sigma_+}(k_+):=\sum_{i\in [k_{+}]} \hat{u}_{\sigma_+(i)}\}_{k_{+} \in [D_+]}$ and $\{\hat{U}_{\sigma_-}(k_-):=\sum_{i\in [k_-]}  \hat{u}_{\sigma_-(i)}\}_{k_{-} \in [D_-]}$ can be done beforehand with time complexity $O(N\log N)$. Thus, the overall running time complexity is $O(N\log N)$. \QEDA

The detailed implementation can be found in Algorithm \ref{FE_GSVMF}. Algorithm \ref{FE_GSVMF} is extremely fast when the values of the continuous variables are fixed, which will be illustrated in \Cref{sec_numerical}. 

\begin{algorithm}
	\caption{Subdata Selection Algorithm for Solving GSVMF with Fixed Continuous Variables}\label{FE_GSVMF}
	\begin{algorithmic}[1]
		\INPUT $\left\{\bfx_i,y_i\right\}_{i\in [N]}$, $\left(\bfw, b ,\bfu\right)$, $\rho>0$, $\lambda>0$
		\State Initialize $k_-=0$, $v_{c1}=0$, $v_{c2}=0$, $k_{1+}=0$, $k_{2+}=0$, and $\D_+=\{i\in[N]:   g_i=+1\},\D_-=\{i\in[N]:   g_i=-1\}$ with $D_+=|\D_+|$, $D_-=|\D_-|$
		\State Sort $\{\hat{u}_i=(u_i-t)/N\}_{i\in \D_+}$ and $\{\hat{u}_i=(u_i-t)/N\}_{i\in \D_-}$ in the ascending order such that $\hat{u}_{\sigma_+(1)}\leq...\leq \hat{u}_{\sigma_+(D_+)}$, $\hat{u}_{\sigma_-(1)}\leq...\leq \hat{u}_{\sigma_-(D_-)}$, respectively
		\State Calculate $\tau_1^*=\arg\max_{i\in [D_+]}\{\hat{u}_{\sigma_+(i)} -\rho/D_+<0\}$ and $\tau_2^*=\arg\max_{i\in [D_-]}\{\hat{u}_{\sigma_-(i)}+\rho/D_+<0\}$
		\State Calculate $\left\{\hat{U}_{\sigma_+}(k_+):=\sum_{i\in [k_{+}]} \hat{u}_{\sigma_+(i)}\right\}_{k_{+} \in [D_+]}$ and $\left\{\hat{U}_{\sigma_-}(k_-):=\sum_{i\in [k_-]}  \hat{u}_{\sigma_-(i)}\right\}_{k_{-} \in [D_-]}$ 
		\Do
		\State \textbf{Case 1} Let $k_{1+}=\min\{\lfloor k_-D_+/D_-\rfloor,\tau_1^*\}$, $v_{c1}=\hat{U}_{\sigma_+}(k_{1+})-\rho k_{1+}/D_+ +\hat{U}_{\sigma_-}(k_{-})+\rho k_{-}/D_-$

		
		\State \textbf{Case 2} Let $k_{2+}=\max\{\lceil k_-D_+/D_-\rceil,\tau_2^*\}$, $v_{c2}=\hat{U}_{\sigma_+}(k_{2+})+\rho k_{2+}/D_+ +\hat{U}_{\sigma_-}(k_{-})-\rho k_{-}/D_-$
		
		
		\State Let $\ell=\arg\min_{j=1,2}\{v_{cj}\}, k_+(k_-)=k_{\ell+}, v_{k_-}=v_{c\ell}$, $k_-=k_-+1$
		\doWhile{$k_-\leq D_-$}  
		\State Let $k_-^*=\arg\min_{k_-=0,\ldots,D_-} \{v_{k_-}\}, k_+^*= k_+(k_-^*), v^*=v_{k_-^*}$, $z_i^*=1$ if $i\in \{\sigma_+(\ell)\}_{\ell\in [k_+^*]}\cup \{\sigma_-(\ell)\}_{\ell\in [k_-^*]}$, and 0, otherwise 
		\OUTPUT $(\bfz^*,v^*)$
	\end{algorithmic}
\end{algorithm}
We remark that (i) the proposed method indeed has the unbiased subdata selection interpretation. The binary variables $\bfz$ indicate fair classification outcomes. Data points with their $z-$ values equal to 1 can be designated as the unbiased selected subdata, using which one can run the classifier to improve classification fairness, as described in the next subsection; and (ii) the subdata selection method can be adapted to solve fairness of black-box classifiers such as convolutional neural network, which will be elaborated in \Cref{sec_extension2}.

\subsection{The Proposed Iterative Refining Strategy (IRS)}
The proposed IRS is inspired by the alternating minimization method \citep{chencluster}, and the latter is a widely-used approach for solving many machine learning problems such as matrix completion problem \citep{lai2017convergence,jain2013low}, and compressive sensing problem \citep{liao2014generalized,abolghasemi2012gradient}.
In this subsection, we develop IRS and study its convergence property and approximation bound. To begin with, we let $H(\bfw, b ,\bfu,\bfz)$ denote the objective function of GSVMF \eqref{GSVMF}.
In IRS, we first construct an initial classification solution by solving SVM \eqref{eq_svm}. At each iteration, we solve GSVMF \eqref{GSVMF} to obtain the optimal $\bfz$ when fixing the continuous variables $(\bfw, b,\bfu)$ by letting their values be equal to the ones obtained in the previous iteration, and then solve GSVMF \eqref{GSVMF} to obtain the optimal $(\bfw, b,\bfu)$ when fixing the values of binary variables $\bfz$. 
We continue this procedure until the improvement is within the tolerance.
The detailed implementation is described in Algorithm \ref{AM_alg}.

\begin{algorithm}
	\caption{IRS for Solving GSVMF \eqref{GSVMF}}\label{AM_alg}
	\begin{algorithmic}[1]
		\INPUT $\left\{\bfx_i,y_i\right\}_{i\in [N]}$, $\rho>0$, $\lambda>0$, $t>0$, tolerance $\delta>0$
		\State Let $\kappa=0$, $(\bfw^{0}, b ^{0})$ be an optimal solution of SVM \eqref{eq_svm}
		\State Let $u_i^0=\max\{1-y_{i} (\bfx_{i}^{\top}\bfw^0+b^0),0\}$ and $z_i^0=\I(u_i^0\leq t)$ for each $i\in [N]$
		\Do
		\State Obtain $\bfz^{\kappa+1}\in \arg\min_{\bfz\in \{0,1\}^N}H(\bfw^{\kappa}, b^{\kappa},\bfu^{\kappa},\bfz)$ using \Cref{FE_GSVMF}
		\State Obtain $\left(\bfw^{\kappa+1}, b ^{\kappa+1},\bfu^{\kappa+1}\right)\in \arg\min_{\bfw, b ,\bfu}\{H(\bfw, b,\bfu,\bfz^{\kappa+1}):\eqref{eq_plane}-\eqref{eq_u}\}$
		\State$\kappa=\kappa+1$
		\doWhile{$H(\bfw^{\kappa-1}, b^{\kappa-1},\bfu^{\kappa-1},\bfz^{\kappa-1})-H(\bfw^{\kappa}, b^{\kappa},\bfu^{\kappa},\bfz^{\kappa})>\delta$} 
		\OUTPUT $\left(\bfw^{\kappa}, b ^{\kappa}, \bfu^{\kappa}, \bfz^{\kappa} \right)$
	\end{algorithmic}
\end{algorithm}
We remark that (i) one can run \Cref{AM_alg} multiple times if they would like to improve the performance by choosing the initial solution randomly; (ii) Step 4 of IRS \Cref{AM_alg} (i.e., the optimization over $\bfz$) can be effectively solved by employing \Cref{FE_GSVMF}; and (iii) the Step 5 of IRS \Cref{AM_alg} (i.e., the optimization over $(\bfw, b,\bfu)$) can be effectively solved using first-order methods or existing packages.


Notice that to select the best tuning parameter $\rho$ from a predetermined list using cross-validation, one might need to run IRS \Cref{AM_alg} multiple times. To accelerate this procedure, we can use warm-starts, i.e., the best solution found when solving the previous GSVMF with a different $\rho$ can be set as an initial solution of the current GSVMF. In practice, this warm-start procedure can significantly improve the convergence of the IRS. 

Next, we observe that the sequence of the objective values $\{H(\bfw^{\kappa}, b^{\kappa},\bfu^{\kappa},\bfz^{\kappa})\}_{\kappa}$ of IRS \Cref{AM_alg} is monotonically non-increasing,  is bounded from below, and is thus convergent. 
\begin{proposition}\label{theorem2}
	The sequence of the objective values $\{H(\bfw^{\kappa}, b^{\kappa} ,\bfu^{\kappa},\bfz^{\kappa})\}_{\kappa}$ of GSVMF \eqref{GSVMF} from IRS \Cref{AM_alg} is monotonically non-increasing and bounded from below, and hence converges.
\end{proposition}

\proof 
In Algorithm \ref{AM_alg}, at $(\kappa+1)$th iteration, given the previous classification decision ${\bm\bfz}^{\kappa}$, the fact that $(\bfw^{\kappa+1}, b^{\kappa+1},\bfu^{\kappa+1}) \in \arg\min_{\bfw, b,\bfu}H(\bfw, b,\bfu,\bfz^{\kappa})$ implies that
\begin{align}
	H(\bfw^{\kappa+1}, b^{\kappa+1},\bfu^{\kappa+1},\bfz^{\kappa})\leq H(\bfw^\kappa, b^\kappa ,\bfu^\kappa,\bfz^\kappa),\label{eq_R_rho_1}
\end{align}
and $\bfz^{\kappa+1}\in \arg\min_{\bfz}H(\bfw^{\kappa+1}, b^{\kappa+1},\bfu^{\kappa+1},\bfz)$ implies that
\begin{align}
	H(\bfw^{\kappa+1}, b^{\kappa+1},\bfu^{\kappa+1},\bfz^{\kappa+1}) \leq H(\bfw^{\kappa+1}, b^{\kappa+1},\bfu^{\kappa+1},\bfz^{\kappa}).\label{eq_R_rho_2}
\end{align}

Summing up \eqref{eq_R_rho_1} and \eqref{eq_R_rho_2} yields $H(\bfw^{\kappa}, b^{\kappa} ,\bfu^{\kappa},\bfz^{\kappa})\geq H(\bfw^{\kappa+1}, b^{\kappa+1},\bfu^{\kappa+1},\bfz^{\kappa+1})$, which implies that the sequence of output objective values $\{H(\bfw^{\kappa}, b^{\kappa} ,\bfu^{\kappa},\bfz^{\kappa})\}_{\kappa}$ of IRS is monotone non-increasing. On the other hand, according to the definition, we have $ H(\bfw^{\kappa}, b^{\kappa} ,\bfu^{\kappa},\bfz^{\kappa})\geq -t$. Hence, the monotone convergence theorem implies that the sequence $\{H(\bfw^{\kappa}, b^{\kappa} ,\bfu^{\kappa},\bfz^{\kappa})\}_{\kappa}$ is indeed convergent.
\QEDA

Additionally, since $\bm{\bfz}^{\kappa}$ is binary for all $\kappa$ and there is only a limited number of binary solutions in the set $\{0,1\}^N$, IRS \Cref{AM_alg} will terminate in a finite number of iterations.

We conclude this section by proving the approximation bound of the output IRS \Cref{AM_alg} solution.
\begin{proposition}\label{prop_conver}Suppose that $(\hat{\bm{w}},\hat{b},\hat{\bm{u}},\hat{\bm{z}})$ denotes a accumulative point of the solution sequence output by IRS \Cref{AM_alg}  and $(\bm{w}^*,b^*,\bm{u}^*,\bm{z}^*)$ denotes an optimal solution of GSVMF \eqref{GSVMF} with the optimal value $v_\rho$. 
	Then we have
	$$v_\rho\leq H(\hat{\bm{w}},\hat{b},\hat{\bm{u}},\hat{\bm{z}})\leq v_\rho+\frac{M_u}{N} | \supp(\bm{z}^*)\Delta \supp(\hat{\bm{z}})|,$$
	where $v_\rho=H(\bm{w}^*,b^*,\bm{u}^*,\bm{z}^*)$ and $M_u$ is defined in \Cref{big_M}.
\end{proposition}
\proof 
Recall that we use $H({\bm{w}},{b},{\bm{u}},{\bm{z}})$ to denote the objective value of GSVMF \eqref{GSVMF}. We observe that 
\begin{subequations}
	\begin{align}\label{b1}
		H(\bm{w}^*,b^*,\bm{u}^*,\bm{z}^*) \leq H(\hat{\bm{w}},\hat{b},\hat{\bm{u}},\hat{\bm{z}}) \leq H(\bm{w}^*,b^*,\bm{u}^*,\hat{\bm{z}}) 
	\end{align}
	where the first inequality is due to feasibility $(\hat{\bm{w}},\hat{b},\hat{\bm{u}},\hat{\bm{z}})$ of  and the second inequality is due to optimality condition of IRS \Cref{AM_alg}. Similarly, we also have 
	\begin{align}\label{b2}
		H(\bm{w}^*,b^*,\bm{u}^*,\bm{z}^*) \leq H(\hat{\bm{w}},\hat{b},\hat{\bm{u}},\hat{\bm{z}}) \leq H(\hat{\bm{w}},\hat{b},\hat{\bm{u}},\bm{z}^*)
	\end{align}
\end{subequations}
By summing up \eqref{b1} and \eqref{b2}, we obtain
\begin{align*}
	2H(\hat{\bm{w}},\hat{b},\hat{\bm{u}},\hat{\bm{z}}) &\leq H(\bm{w}^*,b^*,\bm{u}^*,\hat{\bm{z}}) +H(\hat{\bm{w}},\hat{b},\hat{\bm{u}},\bm{z}^*)\\
	&=\frac{1}{N}\sum_{i\in[N]}(\hat{z}_i-z_i^*)(u_i^*-\hat{u}_i)+H(\hat{\bm{w}},\hat{b},\hat{\bm{u}},\hat{\bm{z}})+H(\bm{w}^*,b^*,\bm{u}^*,\bm{z}^*).
\end{align*}
Thus, we have
\begin{align}
	H(\hat{\bm{w}},\hat{b},\hat{\bm{u}},\hat{\bm{z}}) \leq v_\rho+ \frac{1}{N}\sum_{i\in[N]} (\hat{z}_i-z_i^*)(u_i^*-\hat{u}_i)\leq v_\rho+\frac{1}{N}\sum_{i\in \supp(\bm{z}^*)\Delta \supp(\hat{\bm{z}})}|u_i^*-\hat{u}_i|,
\end{align}
where the second inequality is because of $u_i^*-\hat{u}_i\leq |u_i^*-\hat{u}_i|$ for each $i\in [N]$.
Following the proof of  \Cref{big_M}, without loss of generality, we can restrict $0\leq u_i^*,\hat{u}_i\leq M_u$ for any $i\in [N]$. Thus, we further have 
\begin{align*}
	H(\hat{\bm{w}},\hat{b},\hat{\bm{u}},\hat{\bm{z}}) \leq v_\rho+\frac{M_u}{N} | \supp(\bm{z}^*)\Delta \supp(\hat{\bm{z}})|.
\end{align*}
This completes the proof.
\QEDA

We remark that (i) the solution-quality of IRS \Cref{AM_alg}  depends on the difference between the binary variables (i.e., the subdata selection decisions) $\hat{\bm z}$ and $\bm{z}^*$. This is not surprising since the subdata selection decisions are indeed crucial to GSVMF \eqref{GSVMF} and are responsible for balancing the classification accuracy and fairness; and (ii) Since $M_u$ is independent from the number of data points $N$, thus if $\left| \supp(\bm{z}^*)\Delta \supp(\hat{\bm{z}})\right|=o(N)$, then IRS \Cref{AM_alg} is asymptotically optimal when $N\rightarrow\infty$. This further demonstrates the effectiveness of the proposed solution algorithm.

\section{Variation I: Different Fairness Measures}\label{sec_extension1}
In this section, we demonstrate the following variations of the proposed GSVMF \eqref{GSVMF} with the different fairness measures such as false positive rate (FPR), equal opportunity (a.k.a., false negative rate), and demographic parity. 
All the formulations can be solved by the proposed IRS \Cref{AM_alg} effectively.

\subsection{False Positive Rate Fairness}\label{fpr_fair}
False positive rate (FPR) fairness (see, e.g., \citealt{zafar2019fairness}) is another misclassification measure of disparate mistreatment. A classifier has no FPR disparate mistreatment if the different protected groups with label $y=-1$ have the same misclassification rates. More formally, FPR fairness measure can be defined as follows.
\begin{definition} \label{df_fpr} Given a threshold $t>0$, the FPR fairness measure for the binary classification is
	$$F_{\rm FPR} = \left|\Pr(\tilde{u}\leq t|\tilde g=+1,\tilde{y}=-1)-\Pr(\tilde{u}\leq t|\tilde g=-1,\tilde{y}=-1)\right|,$$
	where $\tilde{u}$ denotes violation margin.
\end{definition} 
In \Cref{df_fpr}, $F_{\rm FPR}$ denotes the absolute difference of the misclassification rates of both groups with label $\tilde{y}=-1$. Using the binary variables $\bfz$, the fairness measure $F_{\rm FPR}$ can be computed by
\begin{subequations}
	\begin{equation} \label{Fair_function_fpr}
		\begin{aligned}
			F_{\rm FPR} &= \left|\Pr(\tilde{u}\leq t|\tilde g=+1,\tilde{y}=-1)-\Pr(\tilde{u}\leq t|\tilde g=-1,\tilde{y}=-1)\right|
			= \left| \sum_{i\in \D_{+-}}\frac{z_i}{D_{+-}}-\sum_{i\in \D_{--}}\frac{z_i}{D_{--}}\right|,
		\end{aligned}
	\end{equation}
	where two sets
	\begin{equation} \label{Fair_fpr_set}
		\begin{aligned}
			&\D_{+-}=\{i\in[N]:   g_i=+1, y_i=-1\}, \D_{--}=\{i\in[N]:   g_i=-1, y_i=-1\},
		\end{aligned}
	\end{equation}
\end{subequations}
represent collections of indices of data points from each protected group with true negative labels $y=-1$, and $D_{+-}=|\D_{+-}|$ is the number of group $g=+1$ data points with label $y=-1$, and $D_{--}=|\D_{--}|$ is the number of group $g=-1$ data points with label $y=-1$.

With this observation, we can replace the OMR fairness measure $\F(\bfz)$ in GSVMF \eqref{GSVMF} by FPR fairness as follows
\begin{equation} \label{GSVMF_fpr}
	\underset{\bfw,b,\bfu,\bfz}\min \left\{ \underbrace{\frac{1}{N}\sum_{i\in[N]} z_{i}(u_i-t)}_{\A(\bfz, t\e-\bfu)}+\lambda \underbrace{\|\bfw\|^{2}_2}_{\R(\bfw,b,\bfu)}+\rho \underbrace{\left| \sum_{i\in \D_{+-}}\frac{z_i}{D_{+-}}-\sum_{i\in \D_{--}}\frac{z_i}{D_{--}}\right|}_{\F(\bfz)}: \eqref{eq_plane}- \eqref{eq_binary_z}\right\}.	
\end{equation}

\subsection{Equal Opportunity Fairness} \label{eo_fair} 
Equal opportunity fairness is a popular fairness measure in which the prediction is independent of the sensitive feature for the data points with positive label (see, e.g., \citealt{olfat2017spectral}). It is worthy of mentioning that equal opportunity fairness is also known as false negative rate fairness. Formally, equal opportunity fairness measure has the following representation.
\begin{definition}\label{def_eo} Given a threshold $t>0$, the equal opportunity fairness measure for the binary classification is
	$$F_{\rm EO}=|\Pr(\tilde{u}\leq t|\tilde g=+1,\tilde{y}=+1)-\Pr(\tilde{u}\leq t|\tilde g=-1,\tilde{y}=+1)|,$$
	where $\tilde{u}$ denotes violation margin.
\end{definition} 
In \Cref{def_eo}, $F_{\rm EO}$ denotes the absolute difference of the probabilities of predicting positive labels for group $g=+1$ and group $g=-1$ with label 
$\tilde{y}=+1$. That is, in this definition, the positive label $y=+1$ is viewed to be more important than its negative counterpart $y=-1$. 
Using the binary variables $\bfz$, the fairness measure $F_{\rm EO}$ can be computed as
\begin{subequations}
	\begin{equation} \label{Fair_function3}
		\begin{aligned}
			F_{\rm EO}&=|\Pr(\tilde{u}\leq t|\tilde g=+1,\tilde{y}=+1)-\Pr(\tilde{u}\leq t|\tilde g=-1,\tilde{y}=+1)|=\left|\sum_{i\in \D_{++}}\frac{z_{i}}{D_{++}}-\sum_{i\in \D_{-+}}\frac{z_{i}}{D_{-+}}\right|,
		\end{aligned}
	\end{equation}
	where two sets
	\begin{equation} \label{Fair_eo_set}
		\begin{aligned}
			&\D_{++}=\{i\in[N]:   g_i=+1, y_i=+1\}, \D_{-+}=\{i\in[N]:   g_i=-1, y_i=+1\}
		\end{aligned}
	\end{equation}
\end{subequations}
represent collections of indices of data points from each protected group with label $y=+1$, and $D_{++}=|\D_{++}|$ is the number of group $g=+1$ data points with label $y=+1$, and $D_{-+}=|\D_{-+}|$ is the number of group $g=-1$ data points with label $y=+1$.


With this observation, we can replace the OMR fairness measure $\F(\bfz)$ in GSVMF \eqref{GSVMF} by equal opportunity fairness as follows
\begin{equation} \label{GSVMF_eo}
	\underset{\bfw,b,\bfu,\bfz}\min \left\{\underbrace{\frac{1}{N} \sum_{i\in[N]} z_{i}(u_i-t)}_{\A(\bfz, t\e-\bfu)}+\lambda \underbrace{\|\bfw\|^{2}_2}_{\R(\bfw,b,\bfu)} +\rho \underbrace{\left|\sum_{i\in \D_{++}}\frac{z_{i}}{D_{++}}-\sum_{i\in \D_{-+}}\frac{z_{i}}{D_{-+}}\right|}_{\F(\bfz)}: \eqref{eq_plane}- \eqref{eq_binary_z} \right\}.
\end{equation}

\subsection{Demographic Parity Fairness}\label{dp_fair}
Demographic parity fairness is one of the common fairness measures in which the prediction is independent of the sensitive feature (see, e.g., \citealt{olfat2017spectral}). The formal definition of demographic parity fairness can be found as below.
\begin{definition}\label{df_dp} Given a threshold $t>0$, the demographic parity fairness measure for the binary classification is
	\begin{align*}
		F_{\rm DP}=&\left|\Pr(\tilde{u}\leq t,\tilde{y}=+1|\tilde g=+1)+\Pr(\tilde{u}>t,\tilde{y}=-1|\tilde g=+1) \right.\\
		&\left. -\Pr(\tilde{u}\leq t,\tilde{y}=+1|\tilde g=-1)-\Pr(\tilde{u}>t,\tilde{y}=-1|\tilde g=-1)\right|,
	\end{align*}
	where $\tilde{u}$ denotes violation margin. 
\end{definition} 
In \Cref{df_dp}, $F_{\rm DP}$ denotes the absolute difference of the probabilities of predicting positive labels for group $g=+1$ and group $g=-1$. 
For the binary classification, we observe that a data point in class $y=+1$ will have positive predicted label if the prediction is correct, and a data point in class $y=-1$ will have positive predicted label if the prediction is incorrect. 
Based on this observation, we can use the binary variables $\bfz$ and their complements $\e-\bfz$ to indicate the positive predicted labels from class $y=+1$ and class $y=-1$ in GSVM \eqref{GSVM}, respectively. Using the binary variables $\bfz$, the fairness measure $F_{\rm DP}$ can be computed by
\begin{equation} \label{Fair_function2}
	\begin{aligned}
		F_{\rm DP}=&\left|\Pr(\tilde{u}\leq t,\tilde{y}=+1|\tilde g=+1)+\Pr(\tilde{u}>t,\tilde{y}=-1|\tilde g=+1) \right.\\
		&\left. -\Pr(\tilde{u}\leq t,\tilde{y}=+1|\tilde g=-1)-\Pr(\tilde{u}>t,\tilde{y}=-1|\tilde g=-1)\right|\\
		=& \left|\frac{\sum_{i\in \D_{++}}z_{i}+\sum_{i\in \D_{+-}}(1-z_{i})}{D_+}-\frac{\sum_{i\in \D_{-+}}z_{i}+\sum_{i\in \D_{--}}(1-z_{i})}{D_-}\right|,
	\end{aligned}
\end{equation}
where the sets $\D_{++}, \D_{+-}, \D_{-+}, \D_{--}$ are defined in \eqref{Fair_fpr_set} and \eqref{Fair_eo_set},
and $D_{+}=|\D_{+}|$ is the number of group $g=+1$ data points, and $D_{-}=|\D_{-}|$ is the number of group $g=-1$ data points.

With this observation, we can replace the OMR fairness measure $\F(\bfz)$ in GSVMF \eqref{GSVMF} by demographic parity fairness as follows
\begin{equation} \label{GSVMF_dp}
	\begin{aligned}
	&	\underset{\bfw,b,\bfu,\bfz}\min \left\{\underbrace{\frac{1}{N}\sum_{i\in[N]} z_{i}(u_i-t)}_{\A(\bfz, t\e-\bfu)}+\lambda\underbrace{\|\bfw\|^{2}_2}_{\R(\bfw,b,\bfu)} \right.\\
		&\left.+\rho \underbrace{\left|\frac{\sum_{i\in \D_{++}}z_{i}+\sum_{i\in \D_{+-}} (1-z_{i})}{D_+}-\frac{\sum_{i\in \D_{-+}}z_{i}+\sum_{i\in \D_{--}}(1-z_{i})}{D_-}\right|}_{\F(\bfz)}: \eqref{eq_plane}- \eqref{eq_binary_z}\right\}.
	\end{aligned}
\end{equation}

\section{Variation II: Different Classifiers and Fairness Measures}\label{sec_extension2}

In this section, we demonstrate the following variations of the proposed framework \eqref{generalized} with the generalized OMR fairness measure: (i) We develop the fair multiclass classification formulation; (ii) We propose the generalized logistic regression formulation with fairness; (iii) We study the generalized kernel SVM formulation with fairness; and (iv) We incorporate the OMR fairness into deep learning models.


\subsection{Fair Multiclass Classification} \label{sec_GMSVM}
If there are multiple classes, GSVMF \eqref{GSVMF} can be modified as the following generalized multiclass SVM with the OMR fairness (GMSVMF)

\begin{subequations}\label{eq_GMSVM}
	\begin{align}
		v_\rho^M:=\underset{\bfw, b ,\bfu,\bfz}{\min} &\underbrace{\frac{1}{N} \sum_{i\in[N]} \sum_{j\in[K]:j \neq y_i}(1-z_{ij})(u_{ij}-t)}_{\A(\bfz, t\e-\bfu)}+\lambda \underbrace{\sum_{j\in[K]}\|\bfw_j\|_2^2}_{\R(\bfw,b,\bfu)}
		+\rho\underbrace{\left|\sum_{i\in \D_+}\frac{z_{iy_i}}{D_+}-\sum_{i\in \D_-}\frac{z_{iy_i}}{D_-}\right|}_{\F(\bfz)},\\
		\text{\rm(GMSVMF)} \ 
		\text{s.t.} \ &\bfw^{\top}_{y_i} \bfx_i+b_{y_i} -\bfw^{\top}_j \bfx_i-b_j \geq 1-u_{ij}, \forall{i\in[N]}, {j\in[K]: j \neq y_i},\label{GMSVM_constr1}\\
		& \sum_{j\in[K]} z_{ij} = 1, \forall{i\in[N]},\label{one_class}\\
		& u_{ij} \geq 0, z_{ij}\in \{0,1\}, \forall{i\in[N]}, {j\in[K]}, \label{GMSVM_constr4}
	\end{align}
\end{subequations}
where $D_+=\sum_{i\in [N]}\mathbbm{1}(g_i=+1)$ and $D_-=\sum_{i\in [N]}\mathbbm{1}(g_i=-1)$ are the numbers of the protected groups, and $K\geq 2$ represents the number of classes. The proposed GMSVMF \eqref{eq_GMSVM} is a special case of the framework \eqref{generalized}. For each $i\in [N],j\in [K]$, let binary variable $z_{ij}=1$ if we predict $i$th data point to be label $j$, and 0, otherwise.  Constraints \eqref{one_class} enforce the classifier to assign exact one label to each instance. 
When $K=2$, GMSVMF \eqref{eq_GMSVM} becomes a binary classification problem with the OMR fairness, where $j=1$ denotes the positive label and $j=2$ denotes the negative label. Note that the notion of OMR fairness measure (i.e., \Cref{df_omr}) is simply extended to the multi-class classification model, which defines the absolute difference of the misclassification rates of both groups.

Note that GMSVMF \eqref{eq_GMSVM} has non-convex terms $\{z_{ij} u_{ij}\}_{i\in [N],j\in [K]: j \neq y_i}$, which can be linearized using the similar McCormick inequalities as \eqref{MIP_constr1}. That is, we introduce new variables $s_{ij}=z_{ij}u_{ij}$ for each $i\in [N],j\in [K]: j \neq y_i$ and then linearize it as
\begin{equation}\label{MIP_constr1_multi} 
	s_{ij}\leq M_u z_{ij}, \quad s_{ij}\leq u_{ij}, \quad s_{ij}\geq 0, \quad s_{ij}\geq u_{ij}-M_u(1-z_{ij}), \forall i\in [N], j\in [K]: j \neq y_i.
\end{equation}
Then we can obtain an equivalent MICP formulation.
\begin{proposition} 
	GMSVMF \eqref{eq_GMSVM} can be recast as the following MICP
	\begin{equation}\label{GMSVM_micp} 
		\begin{aligned}
			v^M_{\rho}=\underset{\bfw, b ,\bfu,\bfz,\bm{s}}{\min} &\left\{\frac{1}{N}\sum_{i\in[N]} \sum_{j\in[K]:j \neq y_i}(u_{ij}-t-s_{ij}+z_{ij}t)+\lambda\sum_{j\in[K]}\|\bfw_j\|_2^2\notag \right.\\
			&\left. +\rho\left|\sum_{i\in \D_+}\frac{z_{iy_i}}{D_+}-\sum_{i\in \D_-}\frac{z_{iy_i}}{D_-}\right|:
			\eqref{GMSVM_constr1}-\eqref{GMSVM_constr4},\eqref{MIP_constr1_multi}\right\}.
		\end{aligned}
	\end{equation}
\end{proposition}

Similar to GSVMF \eqref{GSVMF}, we can decompose the fair multiclass classification problem into two subproblems, \text{\rm GMSVMF-1} and \text{\rm GMSVMF-2} by discussing the sign of the absolute function. 
\begin{proposition} \label{prop2_gmsvm}
	The optimal value of GMSVMF \eqref{eq_GMSVM} is $v_{\rho}^M=\min\{v_{\rho1}^M,v_{\rho2}^M\}$, where $v_{\rho1}^M$ and $v_{\rho2}^M$ are optimal values of the following two optimization problems
	\begin{subequations}
		\begin{equation}\label{GMSVMF_1-constr}
			\begin{aligned}
				v_{\rho1}^M:=\underset{\bfw, b ,\bfu,\bfz}{\text{min}}& \frac{1}{N}\sum_{i\in[N]} \sum_{j\in[K]:j \neq y_i}(1-z_{ij})(u_{ij}-t)+\lambda\sum_{j\in[K]}\|\bfw_j\|_2^2-\rho\sum_{i\in \D_+}\frac{z_{iy_i}}{D_+}+\rho\sum_{i\in \D_-}\frac{z_{iy_i}}{D_-},\\
				\text{\rm (GMSVMF-1)}\  \text{s.t.}&\sum_{i\in \D_+}D_-z_{iy_i}-\sum_{i\in \D_-}D_+z_{iy_i}\leq0,\\
				&\eqref{GMSVM_constr1}-\eqref{GMSVM_constr4};\
			\end{aligned}
		\end{equation}
		\begin{equation}\label{GMSVMF-2_constr}
			\begin{aligned}
				v_{\rho2}^M:=\underset{\bfw, b ,\bfu,\bfz}{\text{min}} &\frac{1}{N}\sum_{i\in[N]}\sum_{j\in[K]:j \neq y_i} (1-z_{ij})(u_{ij}-t)+\lambda\sum_{j\in[K]}\|\bfw_j\|_2^2+\rho\sum_{i\in \D_+}\frac{z_{iy_i}}{D_+}-\rho\sum_{i\in \D_-}\frac{z_{iy_i}}{D_-},\\
				\text{\rm (GMSVMF-2)}\  \text{s.t.}&\sum_{i\in \D_+}D_-z_{iy_i}-\sum_{i\in \D_-}D_+z_{iy_i}\geq0,\\
				&\eqref{GMSVM_constr1}-\eqref{GMSVM_constr4}.
			\end{aligned}
		\end{equation}
	\end{subequations}
\end{proposition}


Next, we show that when fixing the values of the continuous variables $(\bfw,b,\bfu)$, the GMSVMF \eqref{eq_GMSVM}  can be solved efficiently.
\begin{proposition}\label{prop_opt_z_m}
	For any fixed $(\bfw,b,\bfu)$ in GMSVMF \eqref{eq_GMSVM}, optimizing over $\bfz$ can be done in time complexity $O(\max\{NK, N\log N\})$.
\end{proposition}
\proof Due to symmetry, it is sufficient to show that the restricted \text{\rm GMSVMF-1} \eqref{GMSVMF_1-constr} can be solved in $O(N\log{N})$ when the values of $(\bfw,b,\bfu)$ are fixed. We split the proof into three steps.

\noindent \textbf{Step 1. }Observe that to solve the restricted \text{\rm GMSVMF-1} \eqref{GMSVMF_1-constr}, we can first enumerate the value of $\sum_{i\in \D_-}z_{iy_i}$, i.e., we can let $\sum_{i\in \D_-}z_{iy_i}=k_-$ for any integer $k_-\in \{0,1,\ldots, D_-\}$. 

For each $i\in [N]$, by discussing whether $z_{iy_i}=1$ or not, the restricted \text{\rm GMSVMF-1} \eqref{GMSVMF_1-constr} with $\sum_{i\in \D_-}z_{iy_i}=k_-$ is equivalent to solve the following two optimization problems
\begin{subequations}
	\begin{equation}\label{GMSVMF_1-constr_sub1}
		\begin{aligned}
			\underset{\{z_i\}_{i\in \D_+}}{\text{min}} &\left\{\sum_{i\in \D_+}\left[ z_{iy_i}\left[\sum_{j \in [K]:j \neq y_i}(u_{ij}-t)/N-\rho/D_+\right]+ (1-z_{iy_i})\sum_{j \in [K]:j \neq y_i, j \neq j_i^*}(u_{ij}-t)/N\right]:\right.\\
			&\left. \sum_{i\in \D_+}z_{iy_i}\leq \lfloor k_-D_+/D_-\rfloor\right\},
		\end{aligned}
	\end{equation}
	\begin{equation}\label{GMSVMF_1-constr_sub2}
		\begin{aligned}
			\underset{\{z_i\}_{i\in \D_-}}{\text{min}}& \left\{\sum_{i\in \D_-}\left[ z_{iy_i}\left[\sum_{j \in [K]:j \neq y_i}(u_{ij}-t)/N+\rho/D_-\right]+ (1-z_{iy_i})\sum_{j \in [K]:j \neq y_i, j \neq j_i^*}(u_{ij}-t)/N\right]:\right.\\
			&\left.\sum_{i\in \D_-}z_{iy_i}= k_-\right\},
		\end{aligned}
	\end{equation}
\end{subequations}
where $j_i^*\in\arg\max_{j\in[K]:j\neq y_i}\{u_{ij}\}$ denotes the incorrect label that will be predicted for each $i$ if the data point is misclassified. Both problems can be solved efficiently via sorting two lists $\{\hat{c}_{i}=\bar{c}_{i1}-\bar{c}_{i0}\}_{i\in \D_+}$ and $\{\hat{c}_{i}=\bar{c}_{i1}-\bar{c}_{i0}\}_{i\in \D_-}$ 
in the ascending order, where $\bar{c}_{i1}=\sum_{j \in [K]:j \neq y_i}(u_{ij}-t)/N$, and $\bar{c}_{i0}=\sum_{j \in [K]:j \neq y_i, j \neq j_i^*}(u_{ij}-t)/N$ are the costs for correct classification and misclassification, respectively. 

In particular, suppose that $\{\hat{c}_i\}_{i\in \D_+}$ and $\{\hat{c}_i\}_{i\in \D_-}$ are sorted as $\hat{c}_{\sigma_+(1)}\leq\ldots\leq \hat{c}_{\sigma_+(D_+)}$, $\hat{c}_{\sigma_-(1)}\leq...\leq \hat{c}_{\sigma_-(D_-)}$, respectively. Now let $\tau_1^*=\arg\max_{i\in [D_+]}\{\hat{c}_{\sigma_+(i)}-\rho/D_+<0\}$, and $k_{+}=\min\{\lfloor k_-D_+/D_-\rfloor,\tau_1^*\}$. Then let
$z_{ij}^*=1$ if $i\in \{\sigma_+(\ell)\}_{\ell\in [k_+^*]}\cup \{\sigma_-(\ell)\}_{\ell\in [k_-^*]}, j=y_i$, or $i\notin \{\sigma_+(\ell)\}_{\ell\in [k_+^*]}\cup \{\sigma_-(\ell)\}_{\ell\in [k_-^*]},j=j^*_{i}$, and 0, otherwise. Then $\{z_{ij}^*\}_{i\in \D_+},\{z_{ij}^*\}_{i\in \D_-}$ solve the problems \eqref{GMSVMF_1-constr_sub1} and \eqref{GMSVMF_1-constr_sub2}, respectively. 

\noindent \textbf{Step 2. }Then we choose the best $\bfz^*$ which achieves the smallest objective value among the possible ones for all $k_-\in \{0,1,\ldots, D_-\}$.

\noindent \textbf{Step 3. }Note that the sortings of two lists $\{\hat{c}_{i}=\bar{c}_{i1}-\bar{c}_{i0}\}_{i\in \D_+}$ and $\{\hat{c}_{i}=\bar{c}_{i1}-\bar{c}_{i0}\}_{i\in \D_-}$ can be done beforehand with time complexity $O(N\log N)$, while computing $\{j_i^*\}_{i\in [N]}$ takes $O(NK)$ time. Thus, the overall running time complexity is $O(\max\{NK, N\log N\})$. \QEDA

The detailed implementation can be found in Algorithm \ref{FE_GMSVMF}. Finally, we recommend using IRS \Cref{AM_alg} to solve GMSVMF \eqref{eq_GMSVM}, wherein Step 4, using Algorithm \ref{FE_GMSVMF} instead of  \Cref{FE_GSVMF}. 

\begin{algorithm}
	\caption{Subdata Selection Algorithm for Solving GMSVMF \eqref{eq_GMSVM} with Fixed Continuous Variables}\label{FE_GMSVMF}
	\begin{algorithmic}[1]
		\INPUT $\left\{\bfx_i,y_i\right\}_{i\in [N]}$, $\left(\bfw, b ,\bfu\right)$, $\rho>0$, $\lambda>0$
		\State Initialize $k_-=0$, $k_{1+}=0$, $k_{2+}=0$, and $\D_+=\{i\in[N]:   g_i=+1\},\D_-=\{i\in[N]:   g_i=-1\}$ with $D_+=|\D_+|$, $D_-=|\D_-|$
		\State Let $j_i^*=\arg\max_{j\in[K]:j\neq y_i}\{u_{ij}\}$, $\bar{c}_{i1}=\sum_{j \in [K]:j \neq y_i}(u_{ij}-t)/N$, $\bar{c}_{i0}=\sum_{j \in [K]:j \neq y_i, j \neq j_i^*}(u_{ij}-t)/N$ for each $i\in [N]$. Sort $\{\hat{c}_{i}=\bar{c}_{i1}-\bar{c}_{i0}\}_{i\in \D_+}$ and $\{\hat{c}_{i}=\bar{c}_{i1}-\bar{c}_{i0}\}_{i\in \D_-}$ in the ascending order such that $\hat{c}_{\sigma_+(1)}\leq...\leq \hat{c}_{\sigma_+(D_+)}$, $\hat{c}_{\sigma_-(1)}\leq...\leq \hat{c}_{\sigma_-(D_-)}$, respectively
		\State Let $\tau_1^*=\arg\max_{i\in [D_+]}\{\hat{c}_{\sigma_+(i)}-\rho/D_+<0\}$, $\tau_2^*=\arg\max_{i\in [D_-]}\{\hat{c}_{\sigma_-(i)}+\rho/D_+<0\}$, $v_{c1}=\sum_{i \in[N]}\hat{c}_{i0}$ and $v_{c2}=\sum_{i \in[N]}\hat{c}_{i0}$
		\State Calculate $\left\{\hat{C}_{\sigma_+}(k_+):=\sum_{i\in [k_{+}]} \hat{c}_{\sigma_+(i)}\right\}_{k_{+} \in [D_+]}$ and $\left\{\hat{C}_{\sigma_-}(k_-):=\sum_{i\in [k_-]}  \hat{c}_{\sigma_-(i)}\right\}_{k_{-} \in [D_-]}$
		\Do

		\State \textbf{Case 1} Let $k_{1+}=\min\{\lfloor k_-D_+/D_-\rfloor,\tau_1^*\}$, $v_{c1}=\hat{C}_{\sigma_+}(k_{1+})-\rho k_{1+}/D_+ +\hat{C}_{\sigma_-}(k_{-})+\rho k_{-}/D_-$
		
		\State \textbf{Case 2} Let $k_{2+}=\max\{\lceil k_-D_+/D_-\rceil,\tau_2^*\}$, $v_{c2}=\hat{C}_{\sigma_+}(k_{2+})+\rho k_{2+}/D_+ +\hat{C}_{\sigma_-}(k_{-})-\rho k_{-}/D_-$

		\State Let $\ell=\arg\min_{j=1,2}\{v_{cj}\}, k_+(k_-)=k_{\ell+}, v_{k_-}=v_{c\ell}$, $k_-=k_-+1$
		\doWhile{$k_-\leq D_-$}  
		\State Let $k_-^*=\arg\min_{k_-=0,\ldots,D_-} \{v_{k_-}\}, k_+^*= k_+(k_-^*), v^*=v_{k_-^*}$, 
		$z_{ij}^*=1$ if $i\in \{\sigma_+(\ell)\}_{\ell\in [k_+^*]}\cup \{\sigma_-(\ell)\}_{\ell\in [k_-^*]}, j=y_i$, or $i\notin \{\sigma_+(\ell)\}_{\ell\in [k_+^*]}\cup \{\sigma_-(\ell)\}_{\ell\in [k_-^*]},j=j^*_{i}$, and 0, otherwise
		\OUTPUT $(\bfz^*,v^*)$
	\end{algorithmic}
	\vspace{-5pt}
\end{algorithm}


\subsection{Generalized Logistic Regression with Fairness} \label{sec_lr}
Logistic regression is another popular binary classification method, which has been used in many areas such as medical data classification \citep{dreiseitl2002logistic}, susceptibility mapping \citep{ayalew2005application}, and fraud detection \citep{shen2007application}.
Given a dataset $\{(\bfx_i,y_i)\}_{i\in [N]}\subseteq \Re^n\times \{0,1\}$ with $N$ data points, following the similar derivation of GSVMF \eqref{GSVMF}, we propose the generalized logistic regression with fairness (GLRF) as follows 

\begin{subequations} \label{GLRF}
	\begin{align}
		\underset{\bfw,b,\bfz\in \{0,1\}^N}{\min} &\underbrace{\frac{1}{N}\sum_{i\in[N]}z_i \left(t-u_i\right)}_{\A(\bfz, t\e-\bfu)} +\lambda  \underbrace{\|\bfw\|^{2}_2}_{\R(\bfw,b,\bfu)}+\rho \underbrace{\left|\sum_{i\in \D_+}\frac{z_i}{D_+}-\sum_{i\in \D_-}\frac{z_i}{D_-}\right|}_{\F(\bfz)},\\
		\text{s.t. }\quad		&u_i =y_i \log\left(h_{(\bm{w},b)}(\bm{x_i})\right)+(1-y_i)\log\left(1-h_{(\bm{w},b)}(\bm{x_i})\right), \forall{i\in{[N]}},
	\end{align}
\end{subequations}
where $h_{(\bm{w},b)}(\bfx_i)=[1+\exp(-{\bfx}_i^{\top}\bfw-b)]^{-1}$ denotes the sigmoid function. 

We remark that (i) GLRF \eqref{GLRF} can be extended to incorporate other fairness measures in \Cref{sec_extension1}; 
and (ii) GLRF \eqref{GLRF} and its variants can be solved by the IRS \Cref{AM_alg} as well.




\subsection{Generalized Kernel SVM with Fairness}

When the data are not suitable for linear models, kernel SVM (KSVM) can be a better alternative and can map the nonlinear models into a higher dimensional space. It is worth mentioning that the choice of kernel functions and parameters might affect the capability of the classifiers \citep{han2012parameter}.
To incorporate the fairness measure in KSVM, we propose the following GKSVMF formulation:
\begin{equation} \label{ksmvf}
	\begin{aligned}
		\underset{\bfz}{\min}& \left\{K(\bfz,\bfu):=\underbrace{\frac{1}{N} \sum_{i\in[N]}z_i(u_i-t)}_{\A(\bfz, t\e-\bfu)}+\rho\underbrace{\left|\sum_{i\in \D_+}\frac{z_i}{D_+}-\sum_{i\in \D_-}\frac{z_i}{D_-}\right|}_{\F(\bfz)}:z_i\in \{0,1\},\forall{i\in{[N]}}\right\},
	\end{aligned}
\end{equation}
%
where $\bfu$ is the vector of violation margins. Note that the proposed GKSVMF \eqref{ksmvf} is also a special case of the framework \eqref{generalized} with $\R(\bfw,b,\bfu)=0$.


The IRS \Cref{AM_alg} can be adapted to solve GKSVMF \eqref{ksmvf}. We first obtain the initial solution from KSVM. We solve GKSVMF \eqref{ksmvf} to select the unbiased data points with the given prediction margins $\bfu$, and then train KSVM with the selected subdata. We continue this procedure until no improvement or other stopping criteria being invoked. The detailed implementation is described in \Cref{ksvmf_alg}.

\begin{algorithm}
	\caption{IRS for Solving GKSVMF \eqref{ksmvf}}\label{ksvmf_alg}
	\begin{algorithmic}[1]
		\INPUT $\left\{\bfx_i,y_i\right\}_{i\in [N]}$, $\rho>0$, $t>0$
		\State Let $\kappa=0$, $\bfu^{0}$ be an optimal solution of $KSVM(\left\{\bfx_i,y_i\right\}_{i\in [N]})$ 
		\State Let $z_i^0=\I(u_i^{0}\leq t)$ for each $i\in [N]$
		\Do
		\State Obtain $\bfz^{\kappa+1}\in \arg\min_{\bfz\in \{0,1\}^N}K(\bfz, \bfu^{\kappa})$
		\State Obtain $\bfu^{\kappa+1}\in \arg\min_{\bfu}KSVM(\left\{\bfx_i,y_i\right\}_{i\in [N]:z_i^{\kappa+1}=1})$
		\State$\kappa=\kappa+1$
		\doWhile{$K(\bfz^{\kappa-1}, \bfu^{\kappa-1})-K(\bfz^{\kappa}, \bfu^{\kappa})>0$}
		\OUTPUT $\left(\bfz^{\kappa}, \bfu^{\kappa} \right)$
	\end{algorithmic}
\end{algorithm}




We remark that (i) Step 5 in \Cref{ksvmf_alg} can be replaced by other black-box classifiers; (ii) GKSVMF \eqref{ksmvf} can be adapted to incorporate other fairness measures in \Cref{sec_extension1}; and (iii) GKSVMF \eqref{ksmvf} and its variants can be solved by \Cref{ksvmf_alg} as well.



\subsection{Fair Deep Learning} \label{sec_faircnn}

Convolutional neural network (CNN) is a popular method for image classification in deep learning \citep{o2015introduction}. 
The Vanilla CNN (VCNN) might be biased against some protected groups. For example, in medical image diagnosis, VCNN model was reported to show discrimination against the gender \citep{du2020fairness}. 
As far as we are concerned, there is no systematic way to deal with the fairness of VCNN. 

Similar to GKSVMF \eqref{ksmvf}, using the binary variables $\bfz$ to represent the subdata selection decisions, we propose the following formulation for Fair CNN (FCNN):


\begin{equation} \label{cnn}
	\begin{aligned}
		\underset{\bfz\in \{0,1\}^N}{\min} &\left\{C(\bfz,\bfu):=\underbrace{\frac{1}{N}\sum_{i\in[N]}  z_i\left(t-u_i\right)}_{\A(\bfz, t\e-\bfu)}+\rho\underbrace{\left|\sum_{i\in \D_+}\frac{z_i}{D_+}-\sum_{i\in \D_-}\frac{z_i}{D_-}\right|}_{\F(\bfz)}\right\},
	\end{aligned}
\end{equation}
Above, vector $\bfu$ is the output of VCNN, where $u_i$ denotes the probability of correctly predicting the label for data point $i\in [N]$. Different from GKSVMF \eqref{ksmvf}, we use $t \in [0, 1]$ to denote the prediction probability threshold, i.e., for each data point $i\in [N]$, $z_i=1$ if the classification probability $u_i\geq t$, and 0, otherwise. That is, for each data point $i\in [N]$, $z_i=1$ indicates the correct prediction when the classification probability $u_i\geq t$ and $z_i=0$ indicates the incorrect prediction. The proposed FCNN \eqref{cnn} is also a special case of the framework \eqref{generalized} with $\R(\bfw,b,\bfu)=0$.


The IRS \Cref{AM_alg} can be adapted to solve FCNN \eqref{cnn}. We first construct the initial classification solution by solving VCNN. We solve FCNN \eqref{cnn} to select the unbiased subdata, and then train VCNN with the subdata. We continue this procedure until the improvement is within the tolerance. The detailed implementation is described in Algorithm \ref{cnn_alg}.

\begin{algorithm}
	\caption{IRS for Solving FCNN \eqref{cnn}}\label{cnn_alg}
	\begin{algorithmic}[1]
		\INPUT $\left\{\bfx_i,y_i\right\}_{i\in [N]}$, $\rho>0$, $t>0$
		\State Let $\kappa=0$, $\bfu^{0}$ be an optimal solution of $VCNN(\left\{\bfx_i,y_i\right\}_{i\in [N]})$ 
		\State Let $z_i^0=\I(t\leq u_i^{0})$ for each $i\in [N]$
		\Do
		\State Obtain $\bfz^{\kappa+1}\in \arg\min_{\bfz\in \{0,1\}^N}C(\bfz, \bfu^{\kappa})$
		\State Obtain $\bfu^{\kappa+1}\in \arg\min_{\bfu}VCNN(\left\{\bfx_i,y_i\right\}_{i\in [N]:z_i^{\kappa+1}=1})$
		\State$\kappa=\kappa+1$
		\doWhile{$C(\bfz^{\kappa-1}, \bfu^{\kappa-1})-C(\bfz^{\kappa},\bfu^{\kappa})>0$}
		\OUTPUT $\left(\bfz^{\kappa},\bfu^{\kappa} \right)$
	\end{algorithmic}
\end{algorithm}

We remark that (i) Step 5 in \Cref{cnn_alg} can be replaced by other deep learning models or black-box classifiers; (ii) FCNN \eqref{cnn} can be extended to incorporate other fairness measures in \Cref{sec_extension1}; and (iii) FCNN \eqref{cnn} and its variants can be solved effectively by \Cref{cnn_alg}.

\section{Beyond Fairness: Incorporating $F_1$ Score into Binary Classification with Unbalanced Data} \label{sec_unbalanced}

Unbalanced datasets for the binary classification are often refereed to those having significantly uneven instance sizes for the two classes. In an unbalanced dataset, although the majority of instances are from one class, the decision-makers are usually interested in understanding the other class with much fewer instances. Having an unbalanced dataset is common in machine learning. Many problems are inherently unbalanced, such as fraud detection \citep{dal2014learned}, anomaly detection \citep{wang2016probabilistic}, and facial expression recognition \citep{rodriguez2017deep}. 
The unbalanced data can influence the predictive performance considerably since most classification algorithms have been developed with the assumption that instance sizes among different classes are equal. Definitions of different performance metrics and the evaluation of the influence of unbalanced data can be found in \citet{jeni2013facing}.  


Remarkably, in this paper, we use one of the popular performance metrics--$F_1$ score \citep{jeni2013facing} and incorporate it into the classification model to improve the training performance. 
Below is its formal definition.
\begin{definition}
	The $F_1$ score is the harmonic mean of the precision (P) and recall (R): $F_1=2P\cdot R/ (P+R)$, where $P=TP/(TP+FP)$ and $R=TP/(TP+FN)$ and coefficients $TP, FP,FN$ represent numbers of true positive, false positive, and false negative, respectively.	
\end{definition}

Following the same notation from GSVM \eqref{GSVM}, we can also benefit from the binary variables $\bfz$ to calculate true positive (TP), false positive (FP), and false negative (FN). Since $F_1\in [0,1]$ and $F_1$ score higher is better, thus we will penalize $1-F_1$ instead, which is equivalent to
\begin{subequations}
	\begin{align} \label{eq_F_score_set}
		1-F_1&=1-2\frac{P\cdot R}{P+R}=1-\frac{2TP}{2TP+FP+FN} 
		=\frac{\sum_{i\in[N]}(1-z_i)}{N+\sum_{i\in\N_+}z_i-\sum_{i\in\N_-}z_i},
	\end{align}
	where two sets
	\begin{align} \N_+=\{i\in[N]:   y_i=+1\}, \N_-=\{i\in[N]:   y_i=-1\}\label{eq_N_set}
	\end{align}
\end{subequations}
represent collections of indices of data points from each class, respectively, and 
$N_+=|\N_+|$ is the number of data points with positive label, $N_-=|\N_-|$ is the number of data points with negative label.

Next, we show that our framework \eqref{generalized} can be adapted to incorporate $F_1$ score to the binary classification problems with unbalanced data: (i) we study GSVM \eqref{GSVM} with unbalanced data; and (ii) we enhance deep learning with unbalanced data.

\subsection{GSVM with Unbalanced Data}\label{unbalanced}
In this subsection, we will extend GSVM \eqref{GSVM} to incorporate $F_1$ score. Penalizing $1-F_1$ with non-negative penalty parameter $\rho$ into GSVM \eqref{GSVM}, we obtain {\rm{GSVM-$F_1$}}
\begin{equation} \label{GSVMF_1}
	\underset{\bfw, b ,\bfu,\bfz}{\min} \left\{ \underbrace{\sum_{i\in\N_+} \frac{z_i(u_i-t)}{N_+} +\sum_{i\in\N_-}\frac{z_i(u_i-t)}{N_-}}_{\A(\bfz, t\e-\bfu)}+ \lambda \underbrace{\|\bfw\|^{2}_2}_{\R(\bfw,b,\bfu)}+\underbrace{\frac{\rho\sum_{i\in[N]}(1-z_i)}{N+\sum_{i\in\N_+}z_i-\sum_{i\in\N_-}z_i}}_{\F(\bfz)}:\eqref{eq_plane}-\eqref{eq_binary_z}\right\},
\end{equation}
where we normalize the positive and negative instances by their sizes $N_+$ and $N_-$, respectively. This weighting strategy helps increase the prediction accuracy of the minority class, which is widely-used for unbalanced classification \citep{xanthopoulos2014weighted}. Note that the proposed {\rm{GSVM-$F_1$}} \eqref{GSVMF_1} resembles framework \eqref{generalized}, where $\F(\bfz)=1-F_1$. We observe that {\rm{GSVM-$F_1$}} \eqref{GSVMF_1} can be formulated as an MICP using McCormick inequalities \eqref{MIP_constr1} and linearizing the fractional term in the objective. 

\begin{proposition}
	{\rm{GSVM-$F_1$}} \eqref{GSVMF_1} can be formulated as the following MICP
	\begin{equation} \label{GSVMF_1_micp}
		\begin{aligned}
			\underset{\bfw, b ,\bfu,\bfz,\bm s,\eta}{\min}&\sum_{i\in\N_+} \frac{s_i-z_it}{N_+} +\sum_{i\in\N_-}\frac{  s_i-z_it}{N_-}+\lambda\|\bfw\|^{2}_2+\rho\eta,\\
			\text{s.t.} \quad&\eta A\geq \sum_{i\in[N]}(1-z_i)^2,A=N+\sum_{i\in\N_+}z_i-\sum_{i\in\N_-}z_i,\\
			&\eqref{eq_plane}-\eqref{eq_binary_z},\eqref{MIP_constr1}. 
		\end{aligned}
	\end{equation}
\end{proposition}
\proof First of all, linearizing the bilinear terms $\{z_iu_i\}_{i\in [N]}$ using McCormick inequalities, we have
\begin{equation*} 
	\begin{aligned}
		\underset{\bfw, b ,\bfu,\bfz,\bm s}{\min}\left\{\sum_{i\in\N_+} \frac{s_i-z_it}{N_+} +\sum_{i\in\N_-}\frac{  s_i-z_it}{N_-}+\lambda\|\bfw\|^{2}_2+\frac{\rho\sum_{i\in[N]}(1-z_i)}{N+\sum_{i\in\N_+}z_i-\sum_{i\in\N_-}z_i}:\eqref{eq_plane}-\eqref{eq_binary_z},\eqref{MIP_constr1}\right\}.
	\end{aligned}
\end{equation*}
Let us define $A=N+\sum_{i\in\N_+}z_i-\sum_{i\in\N_-}z_i$. Then $\rho(1-F_1)$ in the objective function of {\rm{GSVM-$F_1$}} \eqref{GSVMF_1} can be equivalently represented by $\rho\eta$ such that  $\eta\geq \sum_{i\in[N]}(1-z_i)/A$. Thus, we have the following equivalent formulation of {\rm{GSVM-$F_1$}} as 
\begin{equation*} 
	\begin{aligned}
		\underset{\bfw, b ,\bfu,\bfz,\bm s,\eta}{\min}&\sum_{i\in\N_+} \frac{s_i-z_it}{N_+} +\sum_{i\in\N_-}\frac{  s_i-z_it}{N_-}+\lambda\|\bfw\|^{2}_2+\rho \eta,\\
		\text{s.t.} \quad&\eta \geq \sum_{i\in[N]}(1-z_i)/A,A=N+\sum_{i\in\N_+}z_i-\sum_{i\in\N_-}z_i,\\
		&\eqref{eq_plane}-\eqref{eq_binary_z},\eqref{MIP_constr1}. 
	\end{aligned}
\end{equation*}

According to the equivalence
\begin{align*}
	&\eta \geq \frac{\sum_{i\in[N]}(1-z_i)}{A}=\frac{\sum_{i\in[N]}(1-z_i)^2}{A} \quad
	\Leftrightarrow \quad\eta A \geq \sum_{i\in[N]}(1-z_i)^2,
\end{align*}
we arrive at \eqref{GSVMF_1_micp}. 
\QEDA

For the large-scale instances, we propose to solve {\rm{GSVM-$F_1$}} \eqref{GSVMF_1} using IRS \Cref{AM_alg}, where Step 4 requires a different subdata selection algorithm other than \Cref{FE_GSVMF}. In fact, we can adapt \Cref{FE_GSVMF} to solve {\rm{GSVM-$F_1$}} \eqref{GSVMF_1} with time complexity $O(N\log{N})$ when the continuous variables $(\bfw,b,\bfu)$ are fixed.
\begin{proposition}
	For any fixed $(\bfw,b,\bfu)$ in {\rm{GSVM-$F_1$}} \eqref{GSVMF_1}, optimizing over $\bfz$ can be done in time complexity $O(N\log{N})$.
\end{proposition}
\proof We split the proof into five steps.

\noindent \textbf{Step 1. }Suppose that $\sum_{i\in\N_-}z_i=k_-$ for any integer $k_-\in \{0,1,\ldots, N_-\}$. 
Then the restricted {\rm{GSVM-$F_1$}} \eqref{GSVMF_1} is equivalent to solving the following two optimization problem
\begin{subequations}
	\begin{equation}\label{GSVM-F1-constr_sub1}
		\begin{aligned}
			\{z_i^*\}_{i\in\N_+}\in \arg\underset{\{z_i\}_{i\in\N_+}}{\text{min}} \left\{\sum_{i\in\N_+}\frac{z_i(u_i-t)}{N_+}+\frac{\rho\left(N-\sum_{i\in\N_+}z_i-k_-\right)}{N+\sum_{i\in\N_+}z_i-k_-}\right\},
		\end{aligned}
	\end{equation}
	\begin{equation}\label{GSVM-F1-constr_sub2}
		\begin{aligned}
			\{z_i^*\}_{i\in\N_-}\in \underset{\{z_i\}_{i\in\N_-}}{\text{min}} \left\{\sum_{i\in\N_-}\frac{z_i(u_i-t)}{N_-}:\sum_{i\in\N_-}z_i= k_-\right\},
		\end{aligned}
	\end{equation}
\end{subequations}
where both problems can be solved efficiently via sorting two lists $\{\hat{u}_i:=(u_i-t)/N_+\}_{i\in\N_+}$ and $\{\hat{u}_i:=(u_i-t)/N_-\}_{i\in\N_-}$ in the ascending order. Namely, we suppose that $\{\hat{u}_i\}_{i\in \N_+}$ and $\{\hat{u}_i\}_{i\in \N_-}$ are sorted as $\hat{u}_{\sigma_+(1)}\leq...\leq \hat{u}_{\sigma_+(N_+)}$, $\hat{u}_{\sigma_-(1)}\leq...\leq \hat{u}_{\sigma_-(N_-)}$, respectively.

\noindent \textbf{Step 2. }Observe that the optimization problem \eqref{GSVM-F1-constr_sub2} can be solved by letting $z_{\sigma_-(\ell)}^*=1$ for each $\ell\in [k_-]$, and 0 for each $\ell\in [N_-]\setminus[k_-]$.

\noindent \textbf{Step 3. }Next, let $\tau^*=\arg\max_{\ell\in [N_+]}\{\hat{u}_{\sigma_+(\ell)}<0\}$, where we let $\tau^*=0$ if the maximizer does not exist. In the subproblem  \eqref{GSVM-F1-constr_sub1}, the first term is linear in $\bfz$ and is thus non-increasing in $\{z_{\sigma_+(\ell)}\}_{\ell \in [\tau^*]}$ and is non-decreasing in $\{z_{\sigma_+(\ell)}\}_{\ell \in [N_+]\setminus[\tau^*]}$, the second term is convex and non-increasing over the summation $\sum_{i\in\N_+}z_i:=k_+$. 
Thus, according to monotonicity, we must have $z_{\sigma_+(\ell)}^*=1$ for each $\ell\in [\tau^*]$.

Next, we can use bisection approach to search the best $k_+\in \{\tau^*,\tau^*+1,\ldots,N_+\}$ such that the objective is minimized, denoted by $k_+^*$, i.e., the largest $k_+\geq \tau^*$ such that $$\hat{u}_{\sigma_+(k_+)}+\frac{\rho(N-k_+-k_-)}{N+k_+-k_-}-\frac{\rho(N-(k_+-1)-k_-)}{N+(k_+-1)-k_-}\leq 0.$$
Now let
$z_{\sigma_+(\ell)}^*=1$ if $\ell\in [k_+^*]$ and 0 for each $\ell\in [N_+]\setminus [k_+^*]$. \\
\noindent \textbf{Step 4. }Combining Steps 2 and 3, we have $\{z_i^*\}_{i\in\N_+},\{z_i^*\}_{i\in\N_-}$ solve the subproblems \eqref{GSVM-F1-constr_sub1} and \eqref{GSVM-F1-constr_sub2}, respectively. 
Now we choose the best $\bfz^*$ which achieves the smallest objective value among the possible ones for all $k_-\in \{0,1,\ldots, D_-\}$.\\
\noindent \textbf{Step 5. }Note that the sortings of two lists $\{\hat{u}_i:=(u_i-t)/N_+\}_{i\in\N_+}$ and $\{\hat{u}_i:=(u_i-t)/N_-\}_{i\in\N_-}$ can be done beforehand, and the running time complexity for each bisection is $O(\log(N_+))$. Thus, the overall running time complexity is $O(N\log N)$. \QEDA 

The detailed implementation can be found in Algorithm \ref{alg_GSVMF_1}. 
We remark that (i) {\rm{GSVM-$F_1$}} can be generalized to other binary classification models with $F_1$ score to measure the learning outcomes; and (ii) {\rm{GSVM-$F_1$}} can be solved by the proposed IRS \Cref{AM_alg} when replacing Step 4 with employing \Cref{alg_GSVMF_1} to obtain $\bfz^{t+1}$.

\begin{algorithm}
	\small
	\caption{Algorithm for solving {\rm{GSVM-$F_1$}} with fixed $\left(\bfw, b ,\bfu\right)$}\label{alg_GSVMF_1}
	\begin{algorithmic}[1]
		\INPUT$\left\{\bfx_i,y_i\right\}_{i\in [N]}$, $\left(\bfw, b ,\bfu\right)$, $\rho>0$, $\lambda>0$
		\State Initialize $k_-=0$, $k_+=0$, and $\N_+=\{i\in[N]: y_i=+1\}$, $\N_-=\{i\in[N]: y_i=-1\}$ with $N_+=|\N_+|$, $N_-=|\N_-|$
		
		\State Sort $\{\hat{u}_i=(u_i-t)/N_+\}_{i\in\N_+}$ and $\{\hat{u}_i=(u_i-t)/N_-\}_{i\in\N_-}$ in the ascending order such that $\hat{u}_{\sigma_+(1)}\leq...\leq \hat{u}_{\sigma_+(N_+)}$, $\hat{u}_{\sigma_-(1)}\leq...\leq \hat{u}_{\sigma_-(N_-)}$, respectively
		\State Let $\tau^*=\arg\max_{\ell\in [N_+]}\{\hat{u}_{\sigma_+(\ell)}<0\}$ and let $\tau^*=0$ if the maximizer does not exist 
		\State Calculate $\left\{\hat{U}_{\sigma_+}(k_{b+}):=\sum_{i\in [k_{b+}]} \hat{u}_{\sigma_+(i)}\right\}_{k_{b+} \in [N_+]\setminus[\tau^*-1]}$ and $\left\{\hat{U}_{\sigma_-}(k_-):=\sum_{i\in [k_-]}  \hat{u}_{\sigma_-(i)}\right\}_{k_{-} \in [N_-]}$
		\While{$k_-\leq N_-$}
		
		\State Let $\underline{m}=m$, $k_{b0}=\tau^*$ and $k_{b1}=N_+$ 
		\While{$\tau^*\geq 1$ and $k_{b1}-k_{b0}\geq 1$}
		\State$k_{b+}=\lfloor (k_{b0}+k_{b1})/2\rfloor$
		\State$\Delta=\hat{u}_{\sigma_+(k_{b+})}+\frac{\rho(N-k_{b+}-k_-)}{N+k_{b+}-k_-}-\frac{\rho(N-(k_{b+}-1)-k_-)}{N+(k_{b+}-1)-k_-}$
		\If{$\Delta\leq0$}
		\State$k_{b0}:=k_{b+}$
		\Else
		\State $k_{b1}:=k_{b+}$ 
		\EndIf
		\EndWhile{}
		\State Let $v_{k_-}=\hat{U}_{\sigma_+}(k_{b+})+ \hat{U}_{\sigma_-}(k_{-})+\rho(N-k_{b+}-k_- )/(N +k_{b+}-k_-)$
		\State Let $k_+(k_-)=k_{b+}$ and $k_-=k_-+1$
		\EndWhile{}
		\State Let $k_-^*=\arg\min_{k_-=0,\ldots,N_-} \{v_{k_-}\}, k_+^*= k_+(k_-^*), v^*=v_{k_-^*}$, $z_i^*=1$ if $i\in \{\sigma_+(\ell)\}_{\ell\in [k_+^*]}\cup \{\sigma_-(\ell)\}_{\ell\in [k_-^*]}$, and 0, otherwise 
		\OUTPUT $(\bfz^*,v^*)$
		
	\end{algorithmic}
\end{algorithm}

%
%

\subsection{Deep Learning with Unbalanced Data} \label{sec_bcnn}
Similar to {\rm{GSVM-$F_1$}} \eqref{GSVMF_1}, we propose the following formulation for CNN with $F_1$ score ({\rm{CNN-$F_1$}})
\begin{equation} \label{cnn_unbalanced}
	\underset{\bfz\in \{0,1\}^N}{\min} \left\{  \underbrace{\sum_{i\in\N_+} \frac{z_i\left(t-u_i\right)}{N_+} +\sum_{i\in\N_-}\frac{z_i\left(t-u_i\right)}{N_-}}_{\A(\bfz, t\e-\bfu)}+\underbrace{\frac{\rho\sum_{i\in[N]}(1-z_i)}{N+\sum_{i\in\N_+}z_i-\sum_{i\in\N_-}z_i}}_{\F(\bfz)}\right\},
\end{equation}
where we normalize the positive and negative instances by their sizes $N_+$ and $N_-$, respectively, and $\bfu$ is the solution of VCNN with each entry $u_i$ denoting the probability of correctly predicting the label $y_i$ for each $i\in [N]$.  We see that the proposed {\rm{CNN-$F_1$}} \eqref{cnn_unbalanced} can be viewed as a special case of the framework \eqref{generalized}, where $\R(\bfw,b,\bfu)=0$ and $\F(\bfz)=1-F_1$. 



Note that (i) the proposed IRS \Cref{cnn_alg} can be used to solve {\rm{CNN-$F_1$}} , where we replace Step 4 by employing \Cref{alg_GSVMF_1} to obtain $\bfz^{\kappa+1}$ and in \Cref{alg_GSVMF_1}, and we redefine $\{\hat{u}_i=(t-u_i)/N_+\}_{i\in\N_+}$ and $\{\hat{u}_i=(t-u_i)/N_-\}_{i\in\N_-}$; and (ii) {\rm{CNN-$F_1$}} can be generalized to other deep learning models with $F_1$ score to measure the quality of classification outcomes. 

\section{Numerical Experiments} \label{sec_numerical}
We conduct numerical studies to: (a) demonstrate the performance of \Cref{FE_GSVMF} compared with Gurobi solver, (b) demonstrate the performance of IRS \Cref{AM_alg} compared with Gurobi solver, (c) test whether GSVMF \eqref{GSVMF} can indeed improve classification fairness, (d) compare the performance of GSVMF, GKSVMF, and GLRF, (e) illustrate the performance of GSVMF by comparing our IRS \Cref{AM_alg} for solving GSVMF \eqref{GSVMF_dp} with existing SSVM method proposed by \citet{olfat2017spectral}, and (f) illustrate the performance of fair deep learning and deep learning with unbalanced data. All the instances in this section were executed in Python 3.7 with calls to solver Gurobi (version 9.0 with default settings) on a personal PC with 2.3 GHz Intel Core i9 processor and 16G of memory. Codes of the numerical experiments are available at \url{https://github.com/qingye1/Fair_Classification}.

\subsection{\Cref{FE_GSVMF} Testing}

\noindent\textbf{Experiment 1:} 
In this experiment, we conducted a comparison between the proposed \Cref{FE_GSVMF} and Gurobi solver to demonstrate the effectiveness of \Cref{FE_GSVMF}. We compared the running time for solving GSVMF \eqref{GSVMF} with various problem sizes $N$. 
We varied $N$ over 50, 100, 200, 500, 1000, 1500, 2000, and 5000.
We generated each instance by truncating the wine quality (WQ) dataset \citep{uci} into smaller sizes. 
This dataset contains 12 features and wine quality scores. 
We labeled $y_i=+1$ if a wine has a score of 6 or higher and $y_i=-1$, otherwise. We defined $g_i=+1$ for white wines and $g_i=-1$ for red wines as the sensitive feature. 
We set $b=100$, generated $\bfw$ with each entry being uniformly distributed in the interval $[-10.0, 10.0]$ and generated $\bfu$ with each entry being uniformly distributed in the interval $[0, 10.0]$.
For each instance, we set $t=1$, varied $\rho$ over 0.01, 0.03, 0.05, 0.1, 0.2, 0.5, 0.8, 1.0, 2.0, 3.0, 5.0, and 10.0, and output the average running time in seconds. Note that we set the time limit of Gurobi to be 600 seconds. 

\begin{wrapfigure}{r}{0.48\textwidth}
	\vspace{-28pt}
	\begin{center}
		\includegraphics[width=0.48\textwidth]{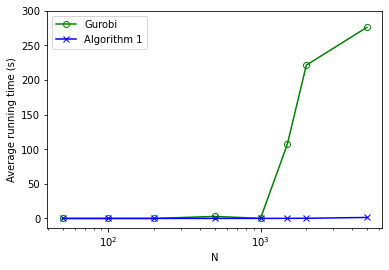}
		\vspace{-28pt}
		\caption{Average Running Time of Gurobi and \Cref{FE_GSVMF} with Different $N$ in Experiment 1. The plot is shown on a logarithmic scale.}\label{alg_test}
	\end{center}
	\vspace{-24pt}
\end{wrapfigure}


Figure \ref{alg_test} illustrates the average running time. We see that \Cref{FE_GSVMF} is always faster than Gurobi. In the first five tests, both \Cref{FE_GSVMF} and Gurobi solve the program within 3 seconds. 
Starting from the size $N=1500$, Gurobi spends a considerably long time on solving GSVMF \eqref{GSVMF}. Since we set the time limit to be 600 seconds, the output time is shorter than the actual time required by Gurobi. Nevertheless, the running time for \Cref{FE_GSVMF} grows much slower than that of Gurobi in the figure. This demonstrates that the proposed \Cref{FE_GSVMF} significantly outperforms Gurobi.

\subsection{IRS \Cref{AM_alg} Testing}

\noindent\textbf{Experiment 2:} In this experiment, we conducted a comparison between IRS \Cref{AM_alg} and Gurobi solver. We compared the objective values and running time for solving GSVMF \eqref{GSVMF} with different $\rho$ and we ran Gurobi to solve MICP \eqref{eq_micp}. We generated the instance by extracting 55 data points from the same WQ dataset in Experiment 1. We set $\lambda=1, t=1$, and varied $\rho$ over 0.01, 0.03, 0.05, 0.1, 0.2, 0.5, 0.8, 1.0, 2.0, 3.0, 5.0, and 10.0.

The running time and objective values are displayed in Table \ref{ram_gurobi}, where opt.val denotes the optimal value of Gurobi and obj.val denotes the objective value of IRS \Cref{AM_alg}. For these cases, Gurobi can solve \eqref{eq_micp} to optimality, and we thus computed the relative optimality gap of IRS \Cref{AM_alg}, denoted by Gap. The running time for the proposed \Cref{AM_alg} is 0.03 seconds for most cases, which is very stable. Gurobi takes a much longer time than IRS. Its running time varies from 9 seconds to 84 seconds, even with $N=55$. The optimality gaps are within 2.5\% for all cases. This demonstrates that IRS can consistently and effectively solve GSVMF \eqref{GSVMF} to near-optimality.

\begin{table}[htbp]
	\centering

	\setlength{\tabcolsep}{5pt}\renewcommand{\arraystretch}{1.0}
	\begin{tabular}{crrrrr}
		\hline
		$N=55, \lambda=1,t=1$  & \multicolumn{2}{c}{Gurobi} & \multicolumn{3}{c}{Proposed IRS}    \\ \cline{2-6}
		$\rho$         & opt.val          & Time (s)    & obj.val        & Gap ($\pct$)     & Time (s)     \\ \hline
		0.01           & -0.6852      & 14    & -0.6827       & 0.4  & 0.03             \\
		0.03           & -0.6829      & 14  & -0.6804         & 0.4    & 0.03           \\
		0.05           & -0.6805      & 13   & -0.6781            & 0.4  & 0.02         \\
		0.1            & -0.6747      & 14  & -0.6722            & 0.4  & 0.02            \\
		0.2            & -0.6631      & 12   & -0.6606          & 0.4    & 0.02         \\
		0.5            & -0.6387      & 9   & -0.6257          & 2.0    & 0.02           \\
		0.8           & -0.6328      & 16  & -0.6275           & 0.8    & 0.03          \\
		1.0             & -0.6309      & 14    & -0.6150        & 2.5      & 0.02       \\
		2.0            & -0.6211      & 29  & -0.6089           & 2.0   & 0.03           \\
		3.0             & -0.6140      & 42  & -0.6075            & 1.1    & 0.03        \\
		5.0            & -0.6112      & 45   & -0.6047         & 1.1     & 0.03          \\
		10.0           & -0.6042      & 84  & -0.5977      & 1.1     & 0.04              \\ \hline
	\end{tabular}%

	\caption{Comparison between Gurobi and the Proposed IRS in Experiment 2}
	\label{ram_gurobi}
	\vspace{-15pt}
\end{table}

\subsection{Subdata Selection Testing} 
\noindent\textbf{Experiment 3:} 
In this experiment, we examined the ability of the proposed GSVMF  \eqref{GSVMF} to improve the fairness by comparing with SVM  \eqref{eq_svm}. We generated 200 data points with two dimensional features $(x_1, x_2)$ besides the sensitive feature $g$. 
The two dimensional features for the data points in different classes and groups were generated from normal distribution with different parameters. Particularly, for the first set of 50 data points with label $y_i=+1$ and sensitive feature $g_i=+1$, their features $(x_1, x_2)$ were generated according to a two-dimensional normal distribution with mean $(3, 4)$ and covariance matrix $\begin{bmatrix}4 \quad 0\\0\quad 9\end{bmatrix}$. For the second set of 50 data points with label $y_i=+1$ and sensitive feature $g_i=-1$, their features $(x_1, x_2)$ were generated according to a two-dimensional normal distribution with mean $(2, 6)$ and covariance matrix $\begin{bmatrix}4\quad 0\\0\quad 9\end{bmatrix}$. For the third set of 50 data points with label $y_i=-1$ and sensitive feature $g_i=+1$, their features $(x_1, x_2)$ were generated according to a two-dimensional normal distribution with mean $(7, 5)$ and covariance matrix $\begin{bmatrix}4\quad 0\\0\quad 9\end{bmatrix}$. For the fourth set of 50 data points with label $y_i=-1$ and sensitive feature $g_i=-1$, their features $(x_1, x_2)$ were generated  according to a two-dimensional normal distribution with mean $(8, 3)$ and covariance matrix $\begin{bmatrix}4\quad 0\\0\quad 9\end{bmatrix}$.

\begin{figure}
	\centering
	\begin{subfigure}[b]{0.49\textwidth}
		\centering
		\includegraphics[width=1.0\linewidth]{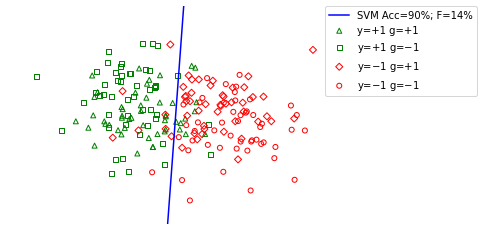}
		\caption{Decision Boundary for SVM }\label{scatter1}
	\end{subfigure}
	\hfill
	\begin{subfigure}[b]{0.49\textwidth}
		\centering
		\includegraphics[width=1.0\linewidth]{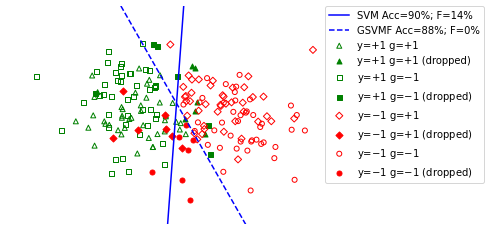}
		\caption{Decision Boundary for GSVMF}\label{scatter2}
	\end{subfigure}
	\caption{A Comparison of SVM and the Proposed GSVMF in Experiment 3.}
	\label{fig:illustrate_gsvmf}
	\vspace{-15pt}
\end{figure}

The SVM results were obtained by solving \eqref{eq_svm} with the best tuned $\lambda=0.5$. Different values of tuning parameters $(t, \lambda, \rho)$ of GSVMF \eqref{GSVMF} were used to optimize the accuracy and fairness level. Particularly, we varied $t$ over $0.1, 0.3, 0.5, 0.7, 0.9, 1.0, 1.5,$ and $2.0$. For $\lambda$, we tried $0, 1/(1000N), 1/(100N), 1/(2N), 1/N,$ $ 2/N,10/N, 100/N,$ and $1000/N$. We varied $\rho$ over $0.01, 0.1,$ $0.2, 0.5, 0.8, 1.0, 2.0, 3.0, 5.0, 10.0,$ and $20.0$. The numerical results are illustrated in \Cref{fig:illustrate_gsvmf}.


\Cref{scatter1} plots the 200 data points and demonstrates the decision boundary for SVM. The accuracy for SVM is 90\% and the OMR fairness is 14\%. Figure \ref{scatter2} illustrates the result of GSVMF, where the best tuning parameters values $(t, \lambda, \rho)$ were selected as $(0.3, 0, 0.2)$. The decision boundary of the proposed GSVMF is to counterclockwise rotates that of SVM by around $30^{\circ}$ to improve the fairness from 14\% to 0\% with only 2\% decrease of accuracy. The data points represented by the filled markers in Figure \ref{scatter2} are the biased ones dropped by the subdata selection method in order to generate a much more fair decision boundary. This demonstrates that incorporating the subdata selection method can largely improve classification fairness. 

%
%



\subsection{Comparisons of Different Classification Models}
\noindent\textbf{Experiment 4 (Comparing Different Binary Classification Methods with OMR Fairness):} In this experiment, we conducted a thorough comparison between GSVMF \eqref{GSVMF}, GKSVMF \eqref{ksmvf}, and GLRF \eqref{GLRF} with OMR fairness in Definition \ref{df_omr} using ProPublica COMPAS dataset \citep{ProPublica} and four real datasets from UCI \citep{uci}.

We preprocessed the data using one-hot encoding and normalization. For COMPAS (CP) dataset, we used the same features as \citet{ProPublica}. We labeled $y_i=+1$ if a person would recidivate within two years and $y_i=-1$, otherwise. We defined $g_i=+1$ for Black and $g_i=-1$ for White. We set gender as the sensitive feature for all the four UCI datasets. Specifically, we defined $g_i=+1$ for male and $g_i=-1$ for female. For default payment (DP) dataset, we labeled $y_i=+1$ for default payment and $y_i=-1$, otherwise. For abalone (AB) dataset, we labeled $y_i=+1$ if the abalone is above the age of 10 years old and $y_i=-1$, otherwise. For Portuguese grade (PG) dataset and math grade (MG) dataset, we labeled $y_i=+1$ for grade above the median and $y_i=-1$, otherwise. 


Different values of tuning parameters $(t, \lambda, \rho)$ of GSVMF \eqref{GSVMF}, GLRF \eqref{GLRF}, and $(t, \rho, kernel)$ of GKSVMF \eqref{ksmvf} were used to obtain the best accuracy and fairness level. 
Particularly, we varied t over 0.1, 0.3, 0.5, 0.7, 0.9, 1.0, 1.5, and 2.0 for GSVMF \eqref{GSVMF} and GKSVMF \eqref{ksmvf}, and  $\log(0.1), \log(0.3), \log(0.5), \log(0.7), \log(0.9),$ $\log(1.0), \log(1.5),$ and $\log(2.0)$ for GLRF \eqref{GLRF}.
For $\lambda$, we tried 0, 1/(1000N), 1/(100N), 1/(2N), 1/N, 2/N, 10/N, 100/N, and 1000/N. We varied $\rho$ over 0.01, 0.1, 0.2, 0.5, 0.8, 1.0, 2.0, 3.0, 5.0, 10.0, and 20.0. For GKSVMF \eqref{GSVMF}, we used RBF and poly kernels. 

We compared GSVMF \eqref{GSVMF}, GKSVMF \eqref{ksmvf}, and GLRF \eqref{GLRF} based on the prediction accuracy (Acc) and fairness (F). Smaller fairness measure (F) is better. The best scenario was selected by the trade-off ratio Acc/F, where a large ratio represents a better trade-off between accuracy and fairness. We used 5-fold cross-validation with a 70/30 training and testing split for all the datasets. GSVMF \eqref{GSVMF}, GKSVMF \eqref{ksmvf}, and GLRF \eqref{GLRF} were solved using the proposed IRS \Cref{AM_alg}, where we used gradient descent method to solve $(\bfw, b, \bfu)$ for GSVMF \eqref{GSVMF} and GLRF \eqref{GLRF}, while GKSVMF \eqref{ksmvf} was solved by \Cref{ksvmf_alg} with a call of python package sklearn.svm.SVC. 

The results displayed in Table \ref{GSVMF-GLRF} are those with the largest Acc/F ratio for each dataset. Their corresponding parameters, accuracy, fairness, and training time are shown in the table. It is seen that GSVMF, GKSVMF, and GLRF have similar prediction accuracy for each dataset. 
For CP, AB, PG, and MG datasets, GSVMF has the best Acc/F ratio. For DP and PG datasets, GKSVMF has the best Acc/F ratio. 
Overall, GKSVMF has shorter training time. Therefore, in practice, we recommend running different models to choose the one having the best fairness given the desired accuracy requirement.

\begin{table}[]
	\centering
	\begin{threeparttable}
		\setlength{\tabcolsep}{2pt}\renewcommand{\arraystretch}{1.0}
		\begin{tabular}{lllllrrrrrrr}
			\hline
			Dataset &Features &$N$ & Methods & Parameters & \multicolumn{3}{c}{Testing} & Time (s)\\ \cline{6-8}
			& & & & & $Acc$ ($\pct$)   & $F$ ($\pct$)   & $Acc/F$  &   \\ \hline
			\multirow{3}{*}{DP} & \multirow{3}{*}{23} &\multirow{3}{*}{30000} 
			&GSVMF   & (0.3, 1/N, 0)\tnote{i} &80.5 &1.8 & 45   & 37.8   \\
			&&  & GKSVMF & (0.5, 0.8, poly)\tnote{ii}  &79.6 &0.1 &796 & 30.3    \\
			&&  & GLRF & ($\log(0.5)$, 0, 0)\tnote{iii}  &81.5  &1.7 &48  &33.1      \\\hline
			
			\multirow{3}{*}{CP} & \multirow{3}{*}{5} &\multirow{3}{*}{5278} 
			&GSVMF   & (2, 1/N, 0) &66.0 &0.02 &3300 &2.5      \\
			&&  & GKSVMF &  (0.7, 2, RBF) &64.0 &0.03 &2133 &1.9    \\
			&&  & GLRF    &($\log(0.3)$, 0, 0) &66.4 &0.06 &1107 &2.5    \\ \hline
			
			\multirow{3}{*}{AB} & \multirow{3}{*}{8} &\multirow{3}{*}{2835} 
			&GSVMF  & (0.5, 1/(1000N), 0.8)&71.8 &0.2 &359 & 2.2      \\
			&&  & GKSVMF &(0.1, 20, poly)&68.6 &0.3 & 229& 1.2    \\
			&&  & GLRF     &($\log(0.1)$, 10/N, 0)&70.4 &2.2 &32 &1.6   \\ \hline
			
			\multirow{3}{*}{PG} & \multirow{3}{*}{32} &\multirow{3}{*}{649} 
			&GSVMF   &  (0.5, 10/N, 0)     &93.3 &0.1 &933   &0.7   \\
			&&  & GKSVMF & (0.5, 0, poly) &93.3 &0.1 &933 &0.2      \\
			&&  & GLRF    & ($\log(0.1)$, 100/N, 0) &94.9 &0.2 &475  & 2.6  \\ \hline
			
			\multirow{3}{*}{MG} & \multirow{3}{*}{31} &\multirow{3}{*}{395} 
			&GSVMF   &  (0.7, 10/N, 0)   &93.3  &0.1  &933  &0.4   \\
			&&  & GKSVMF & (2, 0, RBF) &92.4 &1.8 &51 & 0.1    \\
			&&  & GLRF     & ($\log(0.5)$, 100/N, 0) &91.6 &0.1 &916 &0.8 \\ \hline         
		\end{tabular}%
		\begin{tablenotes}
			\item[i] The 3-tuple represents $(t, \lambda, \rho)$; 
			\item[ii] The 3-tuple represents $(t, \rho, \text{kernel})$;
			\item[iii] The 3-tuple represents $(t, \lambda, \rho)$.
		\end{tablenotes}

		\caption{Performance of GSVMF, GKSVMF, and GLRF with OMR Fairness in Experiment 4}
		\label{GSVMF-GLRF}
	\end{threeparttable}
	\vspace{-10pt}
\end{table}


\noindent\textbf{Experiment 5 (Comparing GSVMF \eqref{GSVMF_dp} and SSVM in \citet{olfat2017spectral} with DP Fairness):} In this experiment, we conducted a comparison between our proposed GSVMF \eqref{GSVMF_dp} and SSVM proposed by \citet{olfat2017spectral} with DP fairness in \Cref{dp_fair} using the same datasets as Experiment 4.


Different values of tuning parameters $(t, \lambda, \rho)$ of GSVMF \eqref{GSVMF_dp} and $(d, \mu)$ of SSVM were used to obtain the best accuracy and fairness level. Particularly, we varied t over 0.1, 0.3, 0.5, 0.7, 0.9, 1.0, 1.5, and 2.0. For $\lambda$, we tried 0, 1/(1000N), 1/(100N), 1/(2N), 1/N, 2/N, 10/N, 100/N, and 1000/N. We varied $\rho$ over 0.01, 0.1, 0.2, 0.5, 0.8, 1.0, 2.0, 3.0, 5.0, 10.0, and 20.0. The values of $(d, \mu)$ were selected to be the same as \citet{olfat2017spectral}.

Similar to {Experiment 4}, we compared GSVMF \eqref{GSVMF_dp} and SSVM based on the prediction accuracy (Acc) and fairness (F). 
The SSVM results were computed using their implementation, which is available at \url{https://github.com/molfat66/FairML}. Both models were tuned using 5-fold cross-validation. 

The results displayed in Table \ref{comparison_table} are those with the largest trade-off ratio for each dataset. The corresponding parameters, accuracy, fairness, and training time are shown in the table. In all the datasets, GSVMF has a considerably larger ratio and shorter training time than SSVM. For some datasets, GSVMF has a slightly lower accuracy than SSVM due to our selection criterion. Nevertheless, the proposed GSVMF's fairness is significantly better than SSVM. Thus, our approach always has a larger Acc/F ratio. Besides, GSVMF can reduce fairness to less than 1\% with good accuracy for the first three datasets. In terms of training time, we see that both methods take a similar amount of time. Therefore, we conclude that using binary variables to incorporate exact fairness measure can indeed significantly improve the fairness. 
\begin{table}[]
	\centering
	\begin{threeparttable}
		\setlength{\tabcolsep}{3pt}\renewcommand{\arraystretch}{1.0}
		\begin{tabular}{lllllrrrrrrr}
			\hline
			Dataset &Features &$N$ & Methods & Parameters& \multicolumn{3}{c}{Testing} & Time (s)\\ \cline{6-8} & & & & & $Acc$ ($\pct$)  & $F$ ($\pct$)   & $Acc/F$  &   \\ \hline
			
			\multirow{2}{*}{DP} & \multirow{2}{*}{23} &\multirow{2}{*}{30000} 
			& SSVM    & (0, 0.01)\tnote{i}      & 71.7     & 2.8    & 26    & 48.5     \\ 
			&& &GSVMF   &(2, 1/(100N), 0.01)\tnote{ii} &82.5 &0.6 & 138 & 39.4      \\ \hline
			
			\multirow{2}{*}{CP} & \multirow{2}{*}{5} &\multirow{2}{*}{5278} 
			& SSVM    &(0.1, 0.01)&58.0 &9.8 &6 &20.7     \\ 
			&& &GSVMF  & (1.5, 1/(2N), 0.5) &64.9 &0.8 &81 & 2.9       \\ \hline

			\multirow{2}{*}{AB} & \multirow{2}{*}{8} &\multirow{2}{*}{2835} 
			& SSVM    &(0.1, 0.01) &73.4 &3.4 &22 &21.8    \\ 
			&&  &GSVMF  &  (0.5, 1/(1000N), 3) &71.8 &0.08 &898 &2.4        \\ \hline

			\multirow{2}{*}{PG} & \multirow{2}{*}{32} &\multirow{2}{*}{649} 
			& SSVM    & (0.1, 1.0)  & 96.7     & 15.9   & 6     &  14.8       \\ 
			&& &GSVMF   &(0.9, 10/N, 0.2)&94.9 &3.6 &26  & 0.8  \\ \hline

			\multirow{2}{*}{MG} & \multirow{2}{*}{31} &\multirow{2}{*}{395} 
			& SSVM    & (0.05, 0.03)  & 96.1     & 9.0   & 11     & 17.6     \\ 
			&& &GSVMF   &(0.5, 10/N, 0)&94.1 &4.3 &22 & 0.5      \\ \hline

		\end{tabular}%
		\begin{tablenotes}
			\item[i] The 2-tuple represents $(d, \mu)$; \item[ii] The 3-tuple represents $(t, \lambda, \rho)$.
		\end{tablenotes}
		\caption{Performance of SSVM and GSVMF with DP Fairness in Experiment 5}
		\label{comparison_table}
	\end{threeparttable}
\end{table}


\subsection{A Comparison between VCNN and FCNN}
\noindent\textbf{Experiment 6:} In this experiment, we conducted a comparison between VCNN (i.e., vanilla CNN) and FCNN \eqref{cnn} using \Cref{cnn_alg}. The datasets used in this experiment were age dataset \citep{cheng2019exploiting}, race dataset \citep{cheng2019exploiting}, gender dataset \citep{zhifei2017cvpr}, and X-ray dataset \citep{xray}. For age dataset, we labeled $y=+1$ for 20s and $y=-1$ for 60s with race of Asian or White as sensitive feature. For race dataset, we labeled $y=+1$ for Asian and $y=-1$ for White with age of 20s or 60s as sensitive feature. For gender dataset, we labeled $y=+1$ for male and $y=-1$ for female with age of 10s or 70s as sensitive feature. For chest X-ray datasets named infiltration and atelectasis, we labeled $y=+1$ for infiltration and atelectasis disease and $y=-1$ for normal with gender of male or female as sensitive feature, respectively. We resized the image data to $50\times50$ pixels and converted to black and white. For both VCNN and FCNN \eqref{cnn}, we chose 20 for epoch, adam for optimizer, and binary cross entropy for loss function and performed training for kernel size (ks) of $3\times3$ and $5\times5$. For FCNN \eqref{cnn}, we varied t over 0.1, 0.3, 0.5, 0.7, 0.9, 0.98, and varied $\rho$ over $0, 1/(10N), 1/(2N), 1/N, 5/N, 10/N$.
In IRS \Cref{cnn_alg}, we solved VCNN using Keras Sequential model in Python deep learning library and then performed IRS procedure for four iterations for the sake of time. The scenario with the best fairness from the four iterations was selected as the output. We ran the IRS \Cref{cnn_alg} for five times and output the average and best results.
We compared the testing accuracy and fairness for VCNN and FCNN \eqref{cnn}. 


Table \ref{cnn_results} displays the testing accuracy (Acc) and fairness (F), kernel size (ks) and the best tuning parameters $(t, \rho)$, and running time for different datasets. Smaller fairness measure (F) is better. It is seen that the proposed FCNN improves both average fairness and accuracy in most scenarios. For example, our approach improves fairness from 12.5\% to 2.7\% with a 4.9\% improvement of accuracy for race dataset when $ks=3\times3$. 
FCNN improves fairness from 12.0\% to 1.2\% with a 1.4\% improvement of accuracy for X-ray infiltration dataset when $ks=5\times5$. 
The best accuracy and fairness for FCNN also outperform VCNN in most datasets. We also observe that due to the iterative procedure in IRS \Cref{cnn_alg}, the training time for our method is longer than that of VCNN. Overall, the proposed FCNN indeed outperforms VCNN both in accuracy and fairness.

\begin{table}[htbp]
	\centering

	\setlength{\tabcolsep}{1.5pt}\renewcommand{\arraystretch}{1.0}
	\begin{tabular}{c|c|c|rr|rr|rr|rr}
		\hline
		\multirow{2}{*}{Dataset} &\multirow{2}{*}{$N$} &\multirow{2}{*}{Methods} &
		\multicolumn{1}{c|}{VCNN} &\multicolumn{1}{c|}{FCNN} &\multicolumn{1}{c|}{VCNN} &
		\multicolumn{1}{c|}{FCNN} &\multicolumn{1}{c|}{VCNN} &\multicolumn{1}{c|}{FCNN} &
		\multicolumn{1}{c|}{VCNN} &\multicolumn{1}{c}{FCNN} \\ \cline{4-11} & & &
		\multicolumn{2}{c|}{Average} &\multicolumn{2}{c|}{Best} &\multicolumn{2}{c|}{Average} &\multicolumn{2}{c}{Best} \\ \hline
		
		\multirow{4}{*}{Age} &\multirow{4}{*}{405} &
		Parameters &\multicolumn{4}{c|}{ks=$3\times3$, t=0.9, $\rho$=1/(2N) } &
		\multicolumn{4}{c}{ks=$5\times5$, t=0.9, $\rho$=1/(2N) } \\ \cline{3-11} 
		&&Testing $Acc (\pct)$ &
		60.2 &63.9 &61.5 &67.2 &57.2 &63.6 &61.5 &65.6 \\ \cline{3-11} 
		& & Testing $F (\pct)$ &
		5.8 &0.3 &2.3 &0.1 &4.5 &0.3 &1.1 &0.1 \\ \cline{3-11} 
		&&Time (s) &
		21 &82 &21 &78 &26 &90 &25 &84 \\ \hline
		
		\multirow{4}{*}{Race} &\multirow{4}{*}{405} &
		Parameters &\multicolumn{4}{c|}{ks=$3\times3$, t=0.9, $\rho$=1/(10N) } &
		\multicolumn{4}{c}{ks=$5\times5$, t=0.98, $\rho$=5/N } \\ \cline{3-11} 
		&&Testing $Acc (\pct)$ &
		53.8 &58.7 &54.9 &61.5 &54.9 &56.1 &57.4 &61.5 \\ \cline{3-11} 
		&& Testing $F (\pct)$ &
		12.5 &2.7 &6.9 &1.1  &10.3 &2.2 &6.9 &0.3 \\ \cline{3-11} 
		&& Time (s) &
		34 &113 &31 &101 &45 &132 &41 &103 \\ \hline
		
		\multirow{4}{*}{Gender} &\multirow{4}{*}{629} &
		Parameters &\multicolumn{4}{c|}{ks=$3\times3$, t=0.98, $\rho$=10/N } &
		\multicolumn{4}{c}{ks=$5\times5$, t=0.98, $\rho$=1/N } \\ \cline{3-11} 
		& & Testing $Acc (\pct)$ &
		66.7 &67.1 &68.1 &69.6 &69.2 &67.7 &72.3 &70.2 \\ \cline{3-11} 
		& & Testing $F (\pct)$ &
		3.5 &0.6 &1.2 &0.1 &5.1 &0.6 &1.2 &0.1 \\ \cline{3-11} 
		& & Time (s) &
		33 &122 &33 &113 &40 &127 &39 &118 \\ \hline
		
		\multirow{4}{*}{\shortstack{X-ray\\Infiltration}} &\multirow{4}{*}{368} &
		Parameters &\multicolumn{4}{c|}{ks=$3\times3$, t=0.5, $\rho$=1/(2N) } &
		\multicolumn{4}{c}{ks=$5\times5$, t=0.5, $\rho$=1/(10N) } \\ \cline{3-11} 
		&&Testing $Acc (\pct)$ &
		67.6 &67.6 &75.7 &72.1 &64.5 &65.9 &69.4 &71.2  \\ \cline{3-11} 
		&& Testing $F (\pct)$ &
		9.8 &0.8 &6.1 &0.5 &12.0 &1.2 &8.4 &0.5  \\ \cline{3-11} 
		&& Time (s) &
		28 &89 &27 &66 &35 &120 &26 &91  \\ \hline
		
		\multirow{4}{*}{\shortstack{X-ray\\Atelectasis}} &\multirow{4}{*}{1122} &
		Parameters &\multicolumn{4}{c|}{ks=$3\times3$, t=0.3, $\rho$=10/N } &
		\multicolumn{4}{c}{ks=$5\times5$, t=0.3, $\rho$=1/(2N) } \\ \cline{3-11} 
		&&Testing $Acc (\pct)$ &
		61.7 &65.6 &64.7 &67.1 &63.1 &63.4 &65.6 &66.2 \\ \cline{3-11} 
		&& Testing $F (\pct)$ &
		2.7 &0.4 &1.5 &0.2 &3.4 &0.6 &1.4 &0.2 \\ \cline{3-11} 
		&& Time (s) &
		36 &129 &34 &119 &43 &143 &42 &139 \\ \hline
	\end{tabular}
	\caption{Performance of FCNN with OMR Fairness in Experiment 6}
	\label{cnn_results}
	\vspace{-10pt}
\end{table}


\subsection{Deep Learning with Unbalanced Datasets}

\noindent\textbf{Experiment 7:} In this experiment, we conducted a comparison between VCNN (i.e., vanilla CNN) and {\rm{CNN-$F_1$}}  \eqref{cnn_unbalanced}. 
We used the same datasets and Keras Sequential model setting as Experiment 6. To generate unbalanced datasets, we reduced the sizes of the datasets for 60s class, Asian class, male class, infiltration class, and atelectasis class to 22\% of age dataset, 22\% of race dataset, 20\% of gender dataset, 24\% of infiltration dataset, and 15\% of atelectasis dataset, respectively. We performed training for kernel size (ks) of $3\times3$ and $5\times5$ and varied $t$ over 0.1, 0.3, 0.5, 0.7, 0.9, 0.98. We also varied $\rho$ over 0, 1/(10N), 1/(2N), 1/N, 5/N, 10/N.
Similar to {Experiment 6}, we solved VCNN using Keras Sequential model in Python deep learning library, while we performed IRS as described in \Cref{sec_bcnn} for four iterations when solving {\rm{CNN-$F_1$}} \eqref{cnn_unbalanced}. The scenario with the best $F_1$ score from the four iterations was selected as the output. We ran the IRS for five times and output the average and best results. 
We compared the testing accuracy and $F_1$ score for VCNN and {\rm{CNN-$F_1$}} \eqref{cnn_unbalanced}.

Table \ref{cnnf1_results} displays the testing accuracy (Acc) and $F_1$ score, kernel size (ks) and the best tuning parameters $(t, \rho)$, and running time for different datasets. Larger $F_1$ score represents better classification result for unbalanced data. It is seen that the proposed {\rm{CNN-$F_1$}} improves both average $F_1$ score and accuracy in most scenarios. 
For example, {\rm{CNN-$F_1$}} improves $F_1$ score from 0.29 to 0.41 with a 7.2\% improvement of accuracy for race dataset when $ks=3\times3$, while it improves $F_1$ score from 0.04 to 0.27 with a 0.3\% improvement of accuracy for X-ray atelectasis dataset when $ks=5\times5$. 
The best accuracy and $F_1$ score for {\rm{CNN-$F_1$}} also outperform VCNN in all scenarios except the atelectasis dataset when $ks=3\times3$.
It is worthy of mentioning that the training time for {\rm{CNN-$F_1$}} is often longer than that of VCNN due to IRS, however, the proposed {\rm{CNN-$F_1$}} can significantly improve the accuracy and $F_1$ score. 

\begin{table}[htbp]
	\centering
	\setlength{\tabcolsep}{0.3pt}\renewcommand{\arraystretch}{1.0}
	\begin{tabular}{c|c|c|rr|rr|rr|rr}
		\hline
		\multirow{2}{*}{Dataset} &\multirow{2}{*}{$N$} &\multirow{2}{*}{Methods} &
		\multicolumn{1}{c|}{VCNN} &\multicolumn{1}{c|}{{\rm{CNN-$F_1$}}} &\multicolumn{1}{c|}{VCNN} &
		\multicolumn{1}{c|}{{\rm{CNN-$F_1$}}} &\multicolumn{1}{c|}{VCNN} &\multicolumn{1}{c|}{{\rm{CNN-$F_1$}}} &
		\multicolumn{1}{c|}{VCNN} &\multicolumn{1}{c}{{\rm{CNN-$F_1$}}} \\ \cline{4-11} & & &
		\multicolumn{2}{c|}{Average} &\multicolumn{2}{c|}{Best} &\multicolumn{2}{c|}{Average} &\multicolumn{2}{c}{Best} \\ \hline
		
		\multirow{4}{*}{Age} &\multirow{4}{*}{274} &
		Parameters &\multicolumn{4}{c|}{ks=$3\times3$, t=0.9, $\rho$=5/N } &
		\multicolumn{4}{c}{ks=$5\times5$, t=0.9, $\rho$=10/N } \\ \cline{3-11} 
		&&Testing $Acc (\pct)$ 
		&73.4 &74.4 &74.6 &80.5 &72.4 &74.8 &74.4 &79.3 \\ \cline{3-11} 
		& & Testing $F_1$ 
		&0.29 &0.32 &0.37 &0.40 &0.33 &0.45 &0.40 &0.49 \\ \cline{3-11} 
		&&Time (s) 
		&20 &79 &19 &75 &26 &104 &26 &101 \\ \hline
		
		\multirow{4}{*}{Race} &\multirow{4}{*}{269} &
		Parameters &\multicolumn{4}{c|}{ks=$3\times3$, t=0.9, $\rho$=1/(10N)} &
		\multicolumn{4}{c}{ks=$5\times5$, t=0.9, $\rho$=0} \\ \cline{3-11} 
		&&Testing $Acc (\pct)$ 
		&72.1 &79.3 &74.1 &81.5 &69.9 &76.3 &72.8 &79.0 \\ \cline{3-11} 
		& & Testing $F_1$ 
		&0.29 &0.41 &0.32 &0.47 &0.24 &0.38 &0.31 &0.41 \\ \cline{3-11} 
		&&Time (s) 
		&20 &75 &20 &73 &24 &98 &24 &94  \\ \hline
		
		\multirow{4}{*}{Gender} &\multirow{4}{*}{391} &
		Parameters &\multicolumn{4}{c|}{ks=$3\times3$, t=0.9, $\rho$=0 } &
		\multicolumn{4}{c}{ks=$5\times5$, t=0.9, $\rho$=1/N } \\ \cline{3-11} 
		& & Testing $Acc (\pct)$ 
		&79.1 &81.4 &79.7 &85.6 &78.8 &83.6 &80.5 &84.7 \\ \cline{3-11} 
		& & Testing $F_1$ 
		&0.31 &0.43 &0.35 &0.45 &0.27 &0.44 &0.34 &0.47 \\ \cline{3-11} 
		& & Time (s) 
		&36 &143 &35 &138 &45 &170 &44 &166 \\ \hline
		
		\multirow{4}{*}{\shortstack{X-ray\\Infiltration}} &\multirow{4}{*}{246} &
		Parameters &\multicolumn{4}{c|}{ks=$3\times3$, t=0.3, $\rho$=0 } &
		\multicolumn{4}{c}{ks=$5\times5$, t=0.1, $\rho$=10/N} \\ \cline{3-11} 
		&&Testing $Acc (\pct)$ 
		&67.8 &71.9 &70.3 &73.0 &68.1 &70.0 &73.0 &75.7 \\ \cline{3-11} 
		& & Testing $F_1$ 
		&0.14 &0.25 &0.32 &0.44 &0.03 &0.37 &0.09 &0.42 \\ \cline{3-11} 
		&&Time (s) 
		&26 &96 &25 &91 &46 &182 &45 &179 \\ \hline
		\multirow{4}{*}{\shortstack{X-ray\\Atelectasis}} &\multirow{4}{*}{653} &
		Parameters &\multicolumn{4}{c|}{ks=$3\times3$, t=0.3, $\rho$=1/N } &
		\multicolumn{4}{c}{ks=$5\times5$, t=0.3, $\rho$=1/N } \\ \cline{3-11} 
		&&Testing $Acc (\pct)$ 
		&82.0 &79.5 &84.2 &82.1 &82.0 &82.3 &82.7 &83.2  \\ \cline{3-11} 
		&& Testing $F_1$ 
		&0.03 &0.28 &0.09 &0.36 &0.04 &0.27 &0.13 &0.33 \\ \cline{3-11} 
		&& Time (s) 
		&60 &236 &59 &231 &64 &253 &63 &246 \\ \hline
	\end{tabular}
	\caption{Performance of {\rm{CNN-$F_1$}} with Unbalanced Data in Experiment 7. Note that larger $F_1$ score implies better classification result.}
	\label{cnnf1_results}
	\vspace{-10pt}
\end{table}

\section{Conclusion}\label{sec_conclusion}

{ We introduced a unified framework for fair classification with unbiased subdata selection procedure and exact fairness representation. The proposed framework is versatile and can be adapted to many classifiers with exact fairness representation. We proposed exact mixed-integer convex programming formulations for the moderate-sized instances, and developed a scalable iterative refining strategy, inspired by the alternating minimization approach, to solve large-scale instances effectively. The numerical study demonstrated that our approach can enhance fairness with little or no loss of prediction accuracy or even improved accuracy. We are working towards incorporating exact fairness into the machine learning models with continuous responses such as fair regression.} 

\section*{Acknowledgment}
We would like to thank Prof. Xinwei Deng from Virginia Tech for bringing up this interesting problem into our attention in Spring 2019. 

\bibliography{ref}


\end{document}